 \documentclass[5p,times,twocolumn]{elsarticle}
\usepackage{amssymb}
\usepackage{acronym}
\usepackage{amsmath}
\usepackage{algorithm}%
\usepackage{algorithmicx}%
\usepackage{algpseudocode}%
\usepackage{graphicx}
\usepackage{subcaption}
\usepackage{hyperref}
\usepackage{xurl}
\usepackage{siunitx}
\usepackage{stfloats} % Helmut added this to allow [b] and [t] options for figures that span two columns
\usepackage{xcolor}

\acrodef{NAS}{Neural Architecture Search}
\acrodef{HPO}{Hyperparameter Optimization}
\acrodef{ML}{Machine Learning}
\acrodef{DL}{Deep Learning}
\acrodef{NN}{Neural Network}
\acrodef{GPU}{Graphics Processing Unit}
\acrodef{CPU}{Central Processing Unit}
\acrodef{HB}{Hyperband}
\acrodef{ASHA}{Asynchronous Successive Halving Algorithm}
\acrodef{RASDA}{Resource-Adaptive Successive Doubling Algorithm}
\acrodef{BOHB}{Bayesian Optimization and HyperBand}
\acrodef{BO}{Bayesian Optimization}
\acrodef{HPC}{High-Performance Computing}
\acrodef{CNN}{Convolutional Neural Network}
\acrodef{CFD}{Computational Fluid Dynamics}
\acrodef{PBT}{Population Based Training}
\acrodef{NCCL}{NVIDIA Collective Communications Library}
\acrodef{SGD}{Stochastic Gradient Descent}
\acrodef{ADAM}{Adaptive Moment Estimation}
\acrodef{GNS}{Gradient Noise Scale}
\acrodef{CV}{Computer Vision}
\acrodef{MSE}{Mean-Squared Error}
\acrodef{AM}{Additive Manufacturing}
\acrodef{GB}{Gigabyte}
\acrodef{TB}{Terrabyte}
\acrodef{SEER}{Sequential Elimination with Elastic Resources}
\acrodef{HEP}{High-Energy Physics}
\acrodef{EO}{Earth Observation}
\acrodef{HEBO}{Heteroscedastic and Evolutionary Bayesian Optimisation}

\journal{Future Generation Computer Systems}

\begin{document}

\begin{frontmatter}


\title{Resource-Adaptive Successive Doubling for Hyperparameter Optimization with Large Datasets on High-Performance Computing Systems}

\author[label1,label2]{Marcel Aach}
\ead{m.aach@fz-juelich.de}
\author[label1]{Rakesh Sarma}
\author[label2]{Helmut Neukirchen}
\author[label1,label2]{Morris Riedel}
\author[label1]{Andreas Lintermann}
% \author{}

\address[label1]{Jülich Supercomputing Centre, Forschungszentrum Jülich GmbH, Germany}
\address[label2]{School of Engineering and Natural Sciences, University of Iceland, Iceland}

\begin{abstract}
The accuracy of \ac{ML} models is highly dependent on the hyperparameters that have to be chosen by the user before the training. However, finding the optimal set of hyperparameters is a complex process, as many different parameter combinations need to be evaluated, and obtaining the accuracy of each combination usually requires a full training run. It is therefore of great interest to reduce the computational runtime of this process. On \ac{HPC} systems, several configurations can be evaluated in parallel to speed up this \ac{HPO}. State-of-the-art \ac{HPO} methods follow a bandit-based approach and build on top of successive halving, where the final performance of a combination is estimated based on a lower than fully trained fidelity performance metric and more promising combinations are assigned more resources over time. Frequently, the number of epochs is treated as a resource, letting more promising combinations train longer. Another option is to use the number of workers as a resource and directly allocate more workers to more promising configurations via data-parallel training. This article proposes a novel \ac{RASDA}, which combines a resource-adaptive successive doubling scheme with the plain \ac{ASHA}. Scalability of this approach is shown on up to 1{,}024 \acp{GPU} on modern \ac{HPC} systems. It is applied to different types of \acp{NN} and trained on large datasets from the \ac{CV}, \ac{CFD}, and \ac{AM} domains, where performing more than one full training run is usually infeasible. Empirical results show that \ac{RASDA} outperforms \ac{ASHA} by a factor of up to $1.9$ with respect to the runtime. At the same time, the solution quality of final \ac{ASHA} models is maintained or even surpassed by the implicit batch size scheduling of \ac{RASDA}. With \ac{RASDA}, systematic \ac{HPO} is applied to a terabyte-scale scientific dataset for the first time in the literature, enabling efficient optimization of complex models on massive scientific data.
\end{abstract}
\acresetall

% \begin{highlights}
% \end{highlights}


\end{frontmatter}
\renewcommand{\thefootnote}{}
\footnotetext{© 2025. This manuscript version is made available under the CC-BY-NC-ND 4.0 license \url{https://creativecommons.org/licenses/by-nc-nd/4.0/}.}
\renewcommand{\thefootnote}{\arabic{footnote}}

\acresetall
\section{Introduction}
\label{sec:intro}
In recent years, the amount of openly available data has drastically increased. This includes datasets from different scientific fields, such as \ac{CV}~\cite{imagenet}, \ac{EO}~\cite{big_earth_net}, \ac{HEP}~\cite{cern_mlpf_data}, \ac{AM}~\cite{BLANC2023100161}, or \ac{CFD}~\cite{Albers:996125}. To analyze these data efficiently and gain novel insights based on hidden correlations, the use of \ac{DL} techniques and \acp{NN} has become essential due to their ability to automatically extract complex patterns. As the prediction quality of these \ac{NN} models is highly dependent on the so-called hyperparameters, which are frequently related to, e.g., the \ac{NN} architecture or the optimizer, systematic \ac{HPO} has become a crucial ingredient of \ac{ML} workflows~\cite{Feurer2019}. However, this search for optimal combinations of hyperparameters is challenging due to often high-dimensional search spaces. Furthermore, the performance of a sample from the search space can only be evaluated with a high degree of confidence after a full model training run. In the case of deep \acp{NN} trained on large datasets, this can become a major hurdle, even with extensive computing resources. Additionally, the search space is often diverse in nature. For example, the search space could be comprised of the learning rate, an optimizer-related parameter represented as floating point number, and the number of layers, an architectural parameter represented as an integer number $l>0$. Categorical values, such as ``type of optimizer'' or ``type of layer'' are also possible. This makes the application of classical, gradient-based optimization methods infeasible. Hyperparameters also change under different models and datasets, making the generalization difficult to assess. One of the state-of-the-art \ac{HPO} methods is the \ac{ASHA}~\cite{ASHA}. It randomly samples multiple combinations, evaluates their performance with a lower training budget and then -- after comparing their performance -- terminates under-performing trials early on. To reduce the time to solution, \ac{ASHA} is frequently executed in parallel, where multiple \ac{NN} configurations (trials) are evaluated at the same time. 

Modern \ac{HPC} systems offer a natural setting for running this kind of workload. They feature accelerators, such as \acp{GPU}, that are ideally suited for efficient \ac{NN} trainings (\textit{fast computation}). Furthermore, these accelerators are connected by an optimized communication network that enables \textit{fast inter-node communication}. While current distributed \ac{HPO} methods, such as \ac{ASHA}, leverage the fast computation capabilities to train different hyperparameter candidates, the communication requirements are usually modest and limited to the exchange of the value of a certain metric, e.g., the current loss on the validation set for the comparison of the performance between trials. 

\begin{figure}
    \centering
    \includegraphics[width=\linewidth]{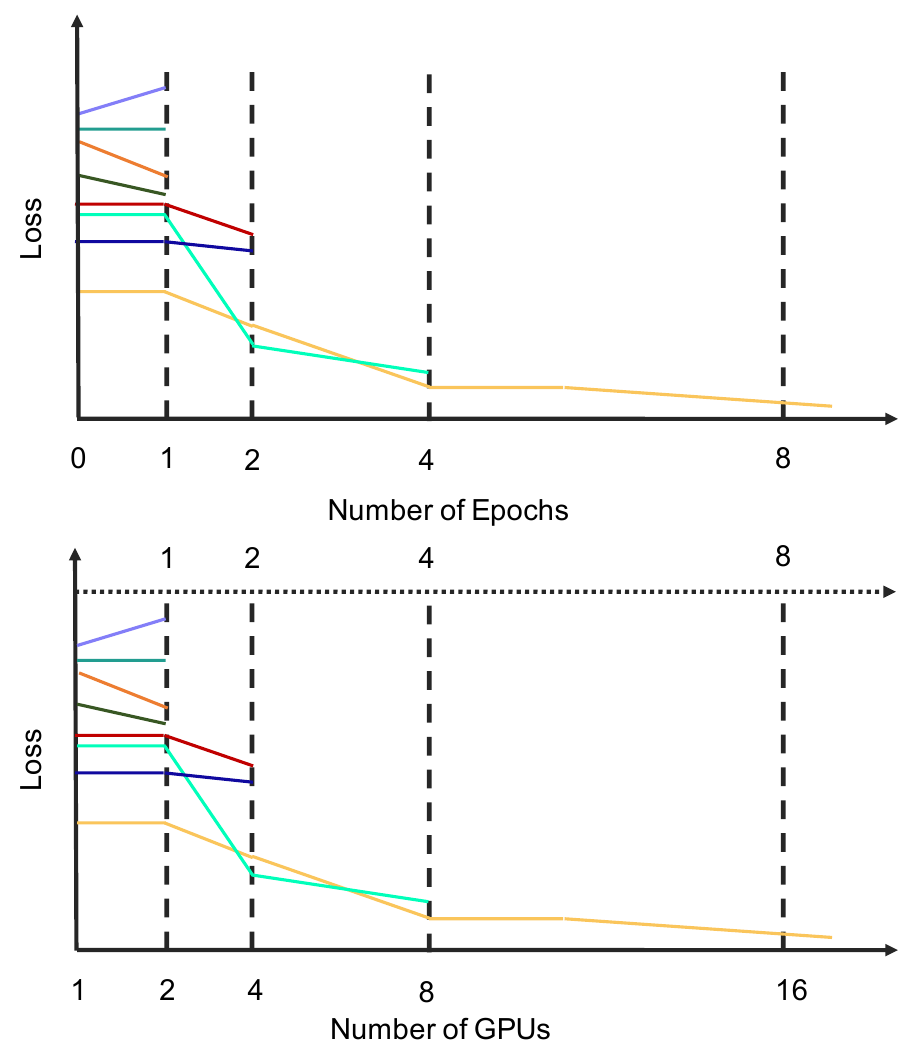}
    \caption{Comparison of successive halving in time (top) and halving in time combined with doubling in space (bottom). Each line corresponds to the learning curve of a single \ac{HPO} combination.}
    \label{fig:time_vs_sapce_halving}
\end{figure}

This work introduces a novel method, the \ac{RASDA}, that leverages \textit{both} \ac{HPC} features to perform \ac{HPO} efficiently at scale. It combines two levels of parallelism: (i) on the \ac{HPO} level, different trials are run in parallel and (ii) on the level of each trial run, the \ac{NN} training is accelerated with data-parallel training. The latter splits the datasets onto multiple \acp{GPU} and performs gradient synchronization after each training step. As these gradients are typically large, they require high-bandwidth communication. \ac{RASDA} then leverages the successive doubling principle, which progressively allocates more resources to more promising hyperparameter combinations, treating the amount of \acp{GPU} that are used for data-parallel training as resources (performing a doubling in space). In contrast, other successive halving techniques, such as the plain \ac{ASHA}, treat the number of epochs during training of a model as resources and thus perform only halving in time, see Fig.~\ref{fig:time_vs_sapce_halving}.

The developed method is suitable for problems that involve large scientific datasets, where due to long training times, even with \ac{HPC} resources it is not feasible to train more than the initially sampled hyperparameter configurations and users are interested in getting the best possible, fully-trained model in the shortest amount of time. Therefore, this study performs an extensive evaluation of \ac{RASDA} on different datasets from the \ac{CV}, \ac{CFD}, and \ac{AM} domains, which are up to 8.3~\ac{TB} in size, to prove its capability to deal with large datasets. These datasets are used to tune the hyperparameters of different types of \acp{NN}, namely a \ac{CNN}, an autoencoder and a transformer. \ac{RASDA} is also benchmarked against the current state-of-the-art successive halving \ac{HPO} method \ac{ASHA}. The new \ac{RASDA} code is openly available on GitHub\footnote{RASDA source: \url{https://github.com/olympiquemarcel/rasda}} for the community, \textcolor{black}{see Tab.~\ref{tab:git_repo_table} for an overview of the repository.}

\textcolor{black}{
\vspace{0.5em} \noindent In summary, the key contributions of \ac{RASDA} are: 
\begin{itemize} 
    \item Combination of successive halving in space with successive doubling in time, allocating more \acp{GPU} to more promising trials.  
    \item Reduces the runtime of more promising hyperparameter trials by leveraging a higher degree of parallelism in data-parallel training  
    \item Leverages the inherent features of \ac{HPC} systems, fast computation for the training, and fast communication for exchange of gradients during distributed training
    \item Outperforms the plain \ac{ASHA} method in runtime and model performance across different domains and on datasets up to 8.3 \ac{TB}. 
\end{itemize} \vspace{0.5em}
}

This article is structured as follows. Section~\ref{sec:related_work} summarizes the related work and highlights the differences to this work. The main details of \ac{RASDA} are presented in Sec.~\ref{sec:algo_presentation}. The application cases are explained in Sec.~\ref{sec:application_cases}, followed by a presentation and discussion of the empirical results of the algorithm in Sec.~\ref{sec:results}. Finally, a summary and outlook are provided in Sec.~\ref{sec:summary}. 

\section{Related Work}\label{sec:related_work}
In \ac{ML}, the performance of a certain model measured by a specific metric, such as the validation error, can be represented by the function $f: \mathcal{X} \rightarrow \mathbb{R}$ where $\mathcal{X}$ denotes the space of possible hyperparameter combinations. The primary goal of \ac{HPO} is to minimize the objective function $f$ by identifying a hyperpara\-meter configuration $x^* \in \mathcal{X}$ such that $x^* \in \arg \min_{x \in \mathcal{X}} f(x)$. Evaluations of the objective function are costly because they typically involve fully training the model for each configuration. To optimize this workflow, several approaches exist. These are either based on approximating $f(x)$ by a lower fidelity estimate, e.g., by the performance after a few training epochs or a model trained on a fraction of the data, or on choosing better hyperparameter configurations to evaluate, e.g., using \ac{BO}). This section summarizes these approaches, i.e. Sec.~\ref{sec:sh} describes the successive halving method, Sec.~\ref{sec:ra_schedulers} and Sec.~\ref{sec:other_algos} summarize resource-adaptive as well as other \ac{HPO} algorithms and Sec.~\ref{sec:data_parallel_dl} introduces the concept of data-parallel training. 

\subsection{Successive Halving}\label{sec:sh}
Successive halving is a variant of Random Search~\cite{random_search}, which uses the fact that most \ac{ML} algorithms are iterative in nature. Intermediate performance results are thus accessible long before the algorithm is fully trained. The problem of finding optimal hyperparameters in a vast search space can then be framed in the context of a multi-armed bandit problem, where each arm represents a hyperparameter combination, and pulling an arm corresponds to training the combination for some iterations~\cite{pmlr-v51-jamieson16}. The goal is to identify the arm that yields the highest reward with the lowest budget possible. To do so in an efficient way, successive halving uniformly allocates an initial budget $B$ to $n_{arms}$ arms and evaluates their performance after a few iterations at a milestone with budget $B / n_{arms}$. It then eliminates the worst-performing half of the arms and promotes the most promising-performing arms by continuing to pull them. Each of these successive halving steps is referred to as a rung. When following this procedure for a few steps from rung to rung, only one arm, i.e., the one with the best performance, remains at the end. \ac{HB}~\cite{hyperband} extends this concept by iterating over different numbers of initial arms $n_{arms}$ (also referred to as brackets) to evaluate. 

However, when performing \ac{HPO} at a larger scale, these methods are sensitive to so-called stragglers. To determine the combinations belonging to the under- and top-per\-for\-ming half, the performance measurement for all combinations needs to be available, which means that faster trials need to wait for the slower ones. \ac{ASHA} addresses this scalability problem by deciding on a rolling basis which trials are worth continuing. When two trials have finished their initial number of iterations, the trial with the better performance is promoted. At the same time, the other trial is paused until the performance of the next completed trial can be juxtaposed. In contrast to \ac{HB}, \ac{ASHA} is mostly performed with only a single bracket, and was evaluated on up to $500$ \acp{GPU} in~\cite{ASHA}.

Another possibility of finding a minimum of the objective function $f$ is to use black-box optimization methods such as \ac{BO}. The idea behind \ac{BO} is to use a probabilistic model of $f$ that is based on data points observed in the past. In the case of \ac{HPO}, this corresponds to finding new promising hyperparameter combinations based on the performance of past combinations. The \ac{BOHB} algorithm~\citep{falkner-icml-18} combines the \ac{BO} process with \ac{HB} for scheduling. To this aim, \ac{HB} is used to choose the number of hyperparameter configurations and their assigned budget, while \ac{BO} is used to choose the hyperparameters by deploying a tree parzen estimator~\citep{NIPS2011_86e8f7ab}.

The mentioned methods have in common that they focus only on identifying the most promising arm and delivering that hyperparameter combination as a result at the end of a run. In contrast, \ac{RASDA} also ensures the full training of the best combination to yield a complete model.

\subsection{Resource-Adaptive Schedulers}\label{sec:ra_schedulers}
Most of the existing successive halving-based \ac{HPO} schedulers treat the number of epochs or training time as resources (also known as fidelity in the literature). It is, however, also possible to treat the spatial amount of computational resources, e.g., the number of \ac{GPU}, used for training a model as a fidelity. A low-fidelity measurement then corresponds to the performance of a \ac{NN} trained with a small number of devices. The most relevant existing \ac{HPO} schedulers that focus on this computational resource-adaptive scheduling are presented in the following.

%hypersched
HyperSched~\cite{hyper_sched} introduces a scheduler to dynamically allocate resources in time and space to the best-performing hyperparameter trials. It thereby not only identifies the most promising model but also trains it -- ideally fully -- by a fixed deadline. The main novelty of the algorithm is its deadline awareness, which means that it schedules fewer new trials as it approaches the deadline. This way, the exploration of new configurations is stopped in favor of deeper exploitation of the running trials. HyperSched is evaluated in~\cite{hyper_sched} on different \ac{CV} benchmarking datasets on up to 32 \acp{GPU} on Cloud computing instances. 

%rubberband
Rubberband~\cite{rubberband} extends HyperSched by leveraging the elasticity of the Cloud for the task of scheduling \ac{HPO} workloads. It takes into account not only the performance of a combination but also the financial costs of a \ac{GPU} hour, with the goal of minimizing the costs of an \ac{HPO} job. Based on the idea of diminishing returns when scaling the training of a single model, the algorithm de-allocates resources (and thus saves costs) from less-promising trials, once a promising trial has been identified. It also creates a resource allocation plan a priori the run to optimize the performance of the single trials that are trained via distributed \ac{DL}.
The resource allocation plan is initialized with an initial burn-in period during which training latencies and scaling performance of trials are measured. 

%seer
\ac{SEER}~\cite{SEER} further takes advantage of the elasticity in the cloud by adaptively allocating and de-allocating compute resources during the \ac{HPO} run. At the same time, it focuses on maximizing the accuracy of trials, in combination with minimizing the total financial cost. Therefore, it limits the amount of workers allocated to the top trials once sub-linear scaling performance sets in. 

Both Rubberband and \ac{SEER} rely heavily on the adaptive allocation and de-allocation of \ac{GPU} instances, which is possible in an elastic cloud setting but not on \ac{HPC} systems, where the amount of \acp{GPU} allocated to the overall \ac{HPO} job is usually static. HyperSched, meanwhile, focuses on maximizing the performance by the deadline. In contrast, the proposed \ac{RASDA} method aims to deliver the best-possible result in the shortest amount of time.

\subsection{Other HPO Algorithms and Libraries}\label{sec:other_algos}
Many other algorithms and libraries for performing \ac{HPO} exist. These include \ac{BO}-based libraries such as Dragonfly~\cite{dragonfly} and SMAC~\cite{smac_3}, allowing the user to select different surrogate models and acquisition functions. Optuna~\cite{optuna} also relies on \ac{BO} and provides automated tracking and visualization of trials. Since parallel computing resources have become increasingly available in recent years, several algorithms have emphasized large-scale, distributed \ac{HPO}: DeepHyper~\cite{deep_hyper} focuses on performing asynchronous \ac{BO} on \ac{HPC} systems and has been applied to several scientific use cases~\cite{cancer_rf, molecular_prediction,LIU2024474}. Distributed evolutionary optimization can be performed with Propulate~\cite{propulate} and \ac{PBT}~\cite{jaderberg2017population}. 

While most of these libraries support multi-fidelity \ac{HPO}, none of them so far supports performing resource-adaptive scheduling of trials, which is, however, supported by \ac{RASDA}. 

\subsection{Data-Parallel Deep Learning}~\label{sec:data_parallel_dl}
Data-parallel training is a technique to reduce the runtime of the training of \ac{DL} models on large datasets by using multiple devices, such as \acp{GPU}. In data-parallel training, the training dataset $\mathcal{D}$ is divided among the number of workers $N$, where each worker is assigned an identical copy of the model to train on a distinct subset of the data $\mathcal{D}_1 \cup \mathcal{D}_2 \cup \ldots \cup \mathcal{D}_N$. Specifically, each worker $i = 1 \dots N$ runs one model forward and backward pass with a predefined number of samples, the local batch size $BS_{local}$, of its subset of data to compute its local gradients $\Delta w_i$ with respect to the model parameters $w$. After the backward pass, these local gradients are aggregated and averaged across all workers by

\begin{equation} \label{eq:gradient_update}
    \Delta w = \frac{1}{N}\sum_{i=1}^N\Delta w_i,
\end{equation}

The averaged global gradient is then used to update the model parameters on all workers every $BS_{global} = BS_{local} \cdot N$ samples~\cite{pytorch_ddp}. 
To remain computationally efficient, each worker needs a sufficient amount of data to run the training, thus $BS_{local}$ needs to be large. At the same time, $BS_{global}$ increases linearly with $N$. When $BS_{global}$ becomes too large, it can impact the generalization performance for two reasons. First, the number of optimizer updates per epoch decreases, as an update is performed every $BS_{global}$ samples. This can be addressed to some extent by scaling the learning rate with the number of devices~\cite{goyal_2017}. This approach is, however, infeasible for an extremely large $BS_{global}$, since in such a case also the learning rate becomes too large. Second, \ac{SGD} with large batch sizes tends to converge to \textit{sharp minima}~\cite{flat_minima} which does not generalize well, see~\cite{keskar2017largebatch} for more details.

\section{Resource-Adaptive Successive Doubling Algorithm}\label{sec:algo_presentation}
This section presents details on \ac{RASDA} in Sec.~\ref{sec:algo_design} and provides an explanation on how issues with large batch size training, cf. Sec.~\ref{sec:data_parallel_dl}, are addressed in Sec.~\ref{sec:large_bs_training}. The performance optimizations
are presented in Sec.~\ref{sec:perf_optimization}, while in Sec.~\ref{sec:comp_existing_hpo} and Sec.~\ref{sec:hardware_dep} the compatibility of \ac{RASDA} with other tools and its dependency on the hardware setup are summarized.

\begin{figure*}[b!] % [b] is understood for figures that span two columns thanks to including package stfloats
    \begin{subfigure}{.5\linewidth}\label{fig:comparison_asha}
    \centering
    \includegraphics[page=2, width=\linewidth]{figures/adaptive_res_pic.pdf}
    \caption{ASHA}
    \end{subfigure}
    \begin{subfigure}{.5\linewidth}\label{fig:comparison_r_asha}
    \centering
    \includegraphics[page=1, width=\linewidth]{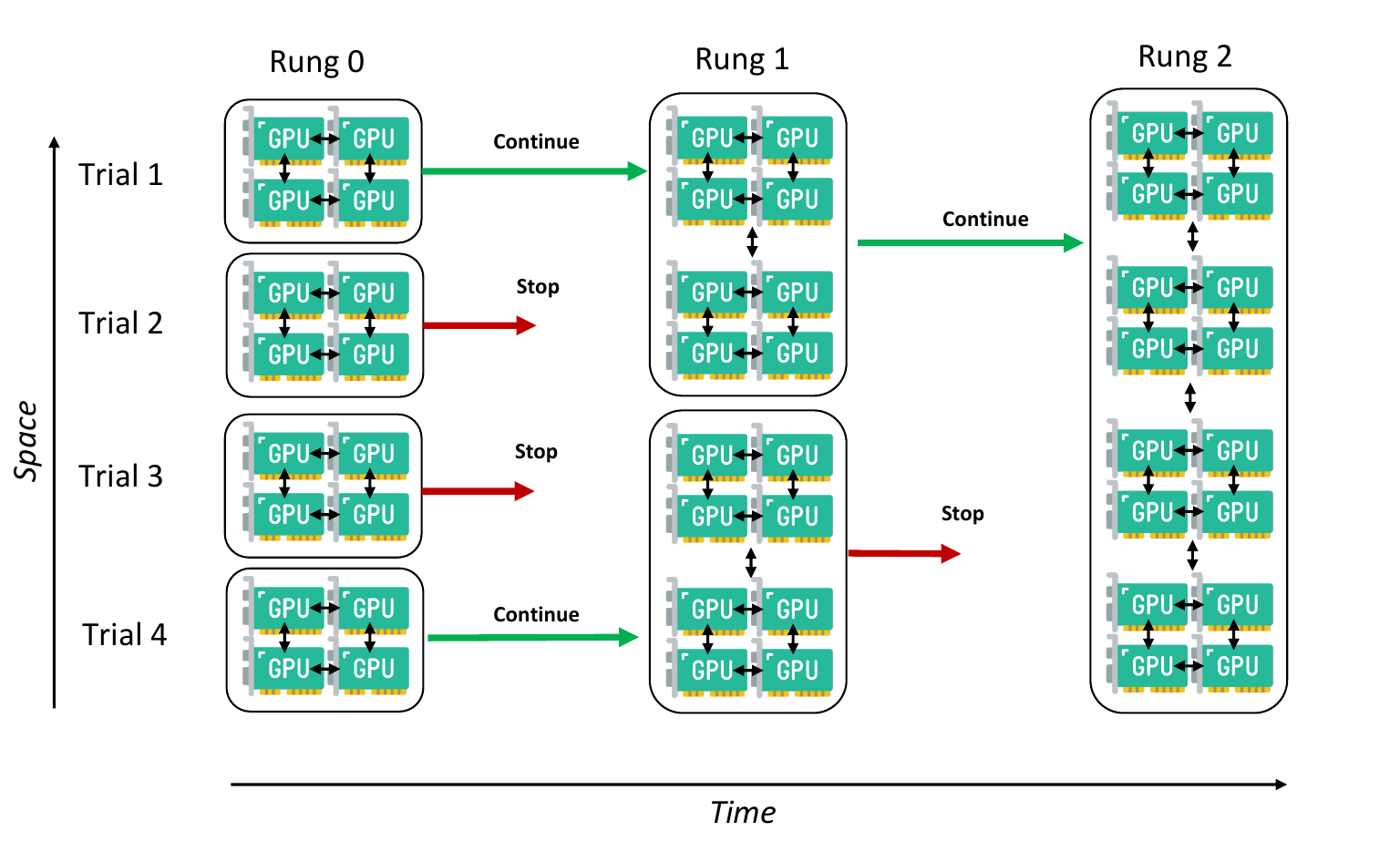}
    \caption{RASDA}
    \end{subfigure}
    \caption{Comparison of (plain) \ac{ASHA}, performing successive halving only in the time domain, and \ac{RASDA}, performing successive halving in the time and successive doubling in the space domain at the same time on an \ac{GPU} cluster. In the \ac{RASDA} case, when a trial is terminated, its workers are allocated to the more promising trials to increase the parallelism of the data-parallel training. Black arrows indicate communication of gradients between \acp{GPU}.} 
    \label{fig:algorithm_comparison}
\end{figure*}

\begin{algorithm*}[b]
\caption{Resource Adaptive Successive Doubling Method}
\renewcommand{\algorithmicrequire}{\textbf{Input:}}
\renewcommand{\algorithmicensure}{\textbf{Output:}}
\begin{algorithmic}[1]
\Require \texttt{trial\_result, base\_resources, sf, milestones}
%\Ensure \texttt{new amount of resources to allocate for the trial}
\If {\texttt{trial\_result["training\_iteration"]} $\in$ \texttt{milestones}}
    \State \texttt{current\_rung} $\gets$ \texttt{milestones.index(trial\_result["training\_iteration"])}
    \State \texttt{new\_resources} $\gets$ \texttt{base\_resources} $\times$ \texttt{sf}$^{\texttt{current\_rung}}$ 
    \State \Return \texttt{new\_resources}
\Else  
    \State \Return \texttt{None}
\EndIf
\end{algorithmic}
\label{alg:resource_successive_halving}
\end{algorithm*}

\subsection{Algorithm Design and Implementation}\label{sec:algo_design}
The main idea of \ac{RASDA} is to combine a successive halving step in the time domain, i.e., train more promising configurations for longer, and a successive doubling step in the spatial domain, i.e., allocate more workers to more promising configurations. This way, when reaching a rung milestone, the worst-performing trials are terminated (halving in time) and the free workers are allocated to the top-performing trials (doubling in space), see Fig.~\ref{fig:algorithm_comparison}. The additional workers are then used to increase the parallelism of the data-parallel training of the configuration, which leads to faster training times. 

For the re-allocation of workers, a second successive doubling routine in addition to the successive halving routine of \ac{ASHA} is used (the resource allocation part is described in Alg.~\ref{alg:resource_successive_halving}): All trials start out with an initial number of workers \\ ($\mathtt{base\_resources}$). When a trial reports a new $\mathtt{trial\_result}$, it is first checked if the current $\mathtt{training\_}$\hspace{0mm}$\mathtt{iteration}$, e.g., the current epoch, corresponds to one of the rung $\mathtt{mile}$$\mathtt{stones}$. At every rung $\mathtt{milestone}$, the (plain) \ac{ASHA} scheduler then reduces the number of running trials by the reduction factor $\mathtt{rf}$. The resources for all trials that are allowed to continue are then increased with the scaling factor $\mathtt{sf}$ by the \ac{RASDA} scheduler, yielding the $\mathtt{new\_resources}$ for the trial (following Alg.~\ref{alg:resource_successive_halving}). If the reduction and scaling factors are equal, i.e.\ $\mathtt{rf}=\mathtt{sf}$, all workers are continuously allocated to a trial. In practice, however, some trials do run faster than others. The advantage of the \ac{ASHA} and \ac{RASDA} scheduler is that they both perform asynchronous halving and doubling, i.e., top-performing trials are promoted to the next rung even if not all trials in the current rung have reached their milestones. This reduces idling times between halving steps. It should be noted that due to this asynchronous execution, the percentage of trials terminated at each milestone can be smaller than $\mathtt{rf}$. As the total number of workers in the system is a constant, the trials that are allowed to continue might need to wait until their new resource requirements are met.

At these rung $\mathtt{milestones}$, two processes occur: the (plain) \ac{ASHA} scheduler reduces the number of running trials by the reduction factor $\mathtt{rf}$, while Alg.\ref{alg:resource_successive_halving} handles the reallocation of GPU resources among the remaining trials.

The total number of rungs and their corresponding milestones in the \ac{RASDA} scheduler are calculated based on the minimum and maximum iterations $\min\_t$ and $\max\_t$, along with the scaling factor $\mathtt{sf}$, as 

\begin{align}
    \mathtt{num\_rungs} &= \left\lfloor \frac{ \log \left( \frac{\max\_t}{\min\_t} \right)}{\log (\mathtt{sf})} \right\rfloor, \\
    \mathtt{rung\_milestones} &= \min\_t \cdot \mathtt{sf}^k  \label{eq:rung_milestones},
\end{align}
with $k = 0, \ldots, \mathtt{num\_rungs}$. This ensures a geometric progression of the milestones, as described for \ac{ASHA} by Li et~al.~\cite{ASHA}.

The algorithm is implemented with Ray Tune~\cite{ray_tune}, an open-source library for performing distributed \ac{HPO}. Ray Tune orchestrates the optimization process by launching a single head node and several worker nodes on an \ac{HPC} cluster. The head node then connects to the worker nodes and starts the trials. During training, the worker nodes report their current status including performance metrics to the head node that makes sche\-duling decisions, such as termination or continuation of new trials. 

Ray Tune already features implementations of several successive halving methods. The implementation of \ac{RASDA} therefore relies on the implementation of \ac{ASHA} that exists already inside of Ray Tune for performing the time-wise successive halving. For the spatial successive doubling, \ac{RASDA} makes use of the \textit{ResourceChangingScheduler} interface\footnote{ResourceChangingScheduler (version 2.8.0): ~\url{https://docs.ray.io/en/latest/tune/api/doc/ray.tune.schedulers.ResourceChangingScheduler.html}}, enabling the modification of resource requirements for trials. At each milestone, the trial is saved, including the current weights of the model. If the decision is made to continue the training, the trial is relaunched with the new resource requirements. It should be noted, that the \ac{ASHA} implementation of Ray Tune has some minor differences to the original algorithm in~\cite{ASHA}. However, empirical evidence shows that these differences do not impact performance.

The data-parallel training part is handled by the PyTorch-DDP library~\cite{pytorch_ddp}, which uses the \ac{NCCL} backend\footnote{\ac{NCCL} backend: \url{https://github.com/NVIDIA/nccl}} for communication and gradient synchronization.

\subsection{Large Batch Training}\label{sec:large_bs_training}
Recall from Sec.~\ref{sec:data_parallel_dl} that scaling the data-parallel training to a large number of devices and increasing $BS_{global}$ can impact the generalization performance of models. The following provides an intuitive explanation of how this issue is addressed by the \ac{RASDA} scheduler.

McCandlish et~al.~\cite{mccandlish2018empirical} empirically studied large batch training for various models: they introduce the \ac{GNS} metric, which serves as a noise-to-signal measure of the training progress. In theory, if the true gradient $G_{true}$ from performing full-batch Gradient Descent without the stochastic component would be available, it would be possible to compute a simple version of the \ac{GNS} by
\begin{align}\label{eq:GNS_definition}
    \mathrm{GNS}_{simple} = \frac{\mathrm{tr}(\Sigma)}{|G_{true}|^2},
\end{align}
where $\Sigma$ is the per-data-sample covariance matrix of $G_{true}$. Essentially, the nominator measures the noise of the gradient, while the denominator measures its magnitude. As the \ac{DL} model converges, the gradient decreases in size, which results in an increase of the \ac{GNS} over training time. McCandlish et~al.~use an approximation to compute the \ac{GNS} based on the estimated stochastic gradient $G_{est}$ and confirm that the \ac{GNS} indeed increases over time. 

Based on the \ac{GNS}, Qiao et~al.~\cite{pollux} introduce the concept of ``statistical efficiency'' of the \ac{DL} training, measuring the amount of training progress made per data sample processed in a batch. The key insight is that when the \ac{GNS} is low, there is no benefit for the learning progress in adding more data samples to the batch (thus increasing $BS_{global}$), as the stochastic gradient $G_{est}$ is a precise approximation of $G_{true}$ already. However, when the \ac{GNS} is high, adding more data samples to the batch reduces the noise and leads to a better gradient approximation. As the \ac{GNS} starts out small and increases over time, this justifies the usage of larger batch sizes during the later part of training. This approach has also been used successfully for \ac{HPO} and scheduling tasks in the past~\cite{pollux, aach_large_bs}.  

Additionally, Smith et~al.~\cite{smith2018dont} find that increasing $BS_{global}$ over time has a similar effect as decaying in the learning rate, which is common practice in \ac{DL} nowadays~\cite{loshchilov2017sgdr}. Based on these findings, the following two insights can be derived:

\begin{itemize}
    \item Training with a small $BS_{global}$ generally helps generalization and is computationally efficient at the beginning of training, in terms of the training progress per processed data sample.
    \item Increasing $BS_{global}$ over time and using a large $BS_{global}$ as the model is converging is computationally efficient as well.
\end{itemize}

This aligns well with the scheduling of the \ac{RASDA} algorithm. In the beginning, the trials train with a small $BS_{global}$, i.e., the number of workers allocated for the data-parallel training is small. As time progresses, $BS_{global}$ increases with each resource doubling step, as more and more workers are allocated to the data-parallel training. The evaluation in Sec.~\ref{sec:results} shows that by leveraging this approach, the generalization capabilities of the final models match or exceed those of models that are continuously trained with a small $BS_{global}$. 

Another crucial point is the correct scaling of the learning rate with the batch size. In the evaluation in Sec.~\ref{sec:results}, the learning rate is scaled linearly with the number of workers, i.e., up to a factor of $8\times$, when using \ac{SGD}~\cite{krizhevsky2014weird}. Furthermore, it follows a square-root scaling rule when using \ac{ADAM}~\cite{malladi2022on}. In the case of re-scaling, the learning rate is not immediately scaled to a larger value. Instead, there is a warm-up over one or two epochs. This re-scaling parameter is included as a hyperparameter in the search space, see Tab.~\ref{tab:search_space}. Thereby, the \ac{HPO} run automatically optimizes towards learning stability. 

\subsection{Performance Optimization}\label{sec:perf_optimization}
To ensure efficient performance, several additional optimizations are made to the trials in the \ac{HPO} loop. This includes selecting the $BS_{local}$ sufficiently large such that it fills the \ac{GPU} memory in addition to the model for each of the applications. As the training datasets have to be loaded by each trial in parallel when performing \ac{HPO}, they are loaded into shared memory when they fit in size. Training datasets that do not fit into shared memory are stored on a partition of the file system with high bandwidth to avoid bottlenecks. For data loading, the native PyTorch data loader as well as the NVIDIA DALI library\footnote{DALI: \url{https://developer.nvidia.com/dali}} are used. 

A preliminary study determined that saving the model weights into a checkpoint too often can lead to bottlenecks~\cite{mixed_precision_hpo}. Therefore, the checkpoint frequency is reduced to every five epochs and the rung milestones of the \ac{ASHA} and \ac{RASDA} scheduler are adjusted accordingly. Ray Tune needs an initial start-up time to launch the head node and all connected worker nodes. As this is the same for \ac{ASHA} and \ac{RASDA}, these timings are excluded from the measurements. 

\textcolor{black}{
\subsection{Compatibility with Existing \ac{HPO} Tools}\label{sec:comp_existing_hpo}
As \ac{RASDA} functions as a pure scheduling tool, it can be integrated with various \ac{HPO} and AutoML frameworks. Since it is already incorporated into the Ray Tune framework via the scheduler interface, it can be seamlessly used within workflows that leverage the scheduling infrastructure of Ray Tune. This includes \ac{BO} tools such as \ac{BOHB}~\cite{falkner-icml-18}, Optuna~\cite{optuna}, and BayesOpt~\cite{bayes_opt}, as well as evolutionary optimization frameworks like HEBO~\cite{hebo}. In such cases, the Bayesian or evolutionary algorithm proposes new hyperparameter candidates to sample from the search space, while \ac{RASDA}s time-wise successive halving determines which trials to terminate at different points in time. Concurrently, its space-wise successive doubling allocates additional \ac{GPU} resources to the trials selected to continue.
}

\subsection{Dependency on Hardware Setup}\label{sec:hardware_dep}
As the main idea of \ac{RASDA} is to perform successive doubling in the spatial domain, the scheduler operates most efficiently on \ac{HPC} systems where the number of \acp{GPU} per node, the total number of allocated nodes, and the number of concurrent trials follow a power-of-two configuration. This setup ensures that \acp{GPU} resources can be reassigned seamlessly across rung milestones, as illustrated in Fig.~\ref{fig:algorithm_comparison}. However, such configurations may not always be available in practice. In scenarios where the number of \acp{GPU} is not a power of two or where the number of trials exceeds the number of available \acp{GPU}, \ac{RASDA} relies on the underlying Ray Tune framework to manage resource scheduling and queuing.
When more trials are submitted than there are \acp{GPU} available, not all trials can be launched simultaneously. In this case, \ac{RASDA} proceeds with its successive doubling routine for the first batch of trials that are scheduled. Remaining trials are queued and executed as resources become available. At the milestones, trials selected to continue are paused if their requested \ac{GPU} allocation cannot be satisfied immediately, and they are resumed once sufficient resources are freed by completed or terminated trials. Although this queuing may introduce some latency, the promoted trials benefit from increased parallelism once resumed, resulting in significantly faster training. The benefit becomes especially pronounced in higher rungs, where larger resource allocations substantially reduce the training time per epoch. As a result, \ac{RASDA} is still expected to deliver well-performing hyperparameter candidates faster than plain \ac{ASHA}. 
Similarly, in cases where the total number of \acp{GPU} is not a power of two, the successive doubling scheme can still be applied, although some adaptation is required. Specifically, the values of $\mathtt{sf}$, $\mathtt{rf}$, $\min\_t$, and $\max\_t$ should be chosen to ensure that the maximum number of \acp{GPU} allocated per trial in the final rung does not exceed the available resources. It is generally advisable to avoid resource fragmentation across nodes, as splitting the \acp{GPU} on a node between multiple trials may reduce the efficiency of data-parallel training.

\section{Application Cases}\label{sec:application_cases}
To assess the proposed \ac{RASDA} scheduler, its performance is evaluated across a range of different tasks from the \ac{CV}, \ac{CFD}, and \ac{AM} domain \textcolor{black}{on seperate training, validation and test dataset splits to avoid overfitting}. The different cases feature various models with different hyperparameters to optimize as well as training datasets of different sizes. The application domains and the set-up of these tasks is described in the following.

\begin{table}[]
\caption{Search space for the experiments, comprised of several optimizer-related and architectural parameters. Superscripts indicate which hyperparameters are used as search space for which applications: \ac{CV}, \ac{CFD}, \ac{AM}. The ''re-scaling warm-up´´ parameter handles the gradual increase of the learning rate when the number of devices and with it $BS_{global}$ is increased.}
\begin{tabular}{l|l|l}
\textbf{Hyperparameter}          & \textbf{Type}  & \textbf{Range}                \\
\hline
\vspace{-10pt} &  &\\
Learning rate$^{\text{CV, CFD, AM}}$ & float & log{[}1e-5, 1{]}    \\
Weight decay$^{\text{CV, CFD, AM}}$      & float   & log${(0, 1e-1]}$   \\
Initial warm-up$^{\text{CV, CFD, AM}}$     & int   & [1,2,3,4,5]   \\
Optimizer$^{\text{CV, CFD, AM}}$     & cat   & ["sgd", "adam"]   \\
Layer initialization$^{\text{CV, CFD}}$ & cat & ["kaiming"~[\cite{10.1109/ICCV.2015.123}],  \\
& & "xavier"\cite{pmlr-v9-glorot10a}] \\
Activation function$^{\text{CV, CFD}}$ & cat & ["ReLU", \\ 
& & "LeakyReLU", \\
& & "SELU", "Tanh", \\
& & "Sigmoid"] \\
Convolution kernel size$^{\text{CV, CFD}}$ & int & [5,7,9] \\
Re-scaling warm-up$^{\text{CFD, AM}}$     & int   & [1,2]   \\
Patch size$^{\text{AM}}$ & int & [2, 4] \\
Depth$^{\text{AM}}$ & int & [1, 2, 4] \\
Number of attention heads$^{\text{AM}}$ & int & [3, 6, 12, 24] \\
MLP ratio$^{\text{AM}}$ & float & [1., 2., 3., 4.]\\
\end{tabular}
\label{tab:search_space}
\end{table}

\subsection{Computer Vision}
For the \ac{CV} domain, the hyperparameters of a ResNet50~\cite{he2016} trained on the ImageNet dataset~\cite{imagenet} are optimized, as this is still one of the most important reference benchmarks~\cite{MLSYS2020_411e39b1}. The ImageNet dataset contains 1{,}281{,}167 training images and 50{,}000 validation images divided into 1{,}000 object classes. In TFRecord file format, the dataset is approximately $146$ \ac{GB} in size. 

The ResNet follows a basic \ac{CNN} architecture with multiple residual connections between layers. The \ac{HPO} search space for the ResNet includes several architectural hyperparameters, e.g., the type of activation functions or size of the input convolution kernel, as well as optimizer-related parameters, such as the learning rate or weight decay, see Tab.~\ref{tab:search_space} for an exhaustive list. 

All models are trained for $\min\_t = 5$ to $\max\_t$ $= 40$ epochs and a reduction and scaling factor $\mathtt{sf}=\mathtt{rf}=2$ is chosen for the schedulers. Following Eq.~\eqref{eq:rung_milestones}, this results in rung milestones at epochs $5, 10, 20,$ and $40$. 

As classification accuracy score, the percentages of the correctly classified training, validation, and test images are computed. 

\subsection{Additive Manufacturing}
The \ac{AM} dataset is taken from the RAISE-LPBF benchmarking dataset~\cite{BLANC2023100161}, which includes a selection of high-speed video recordings at 20{,}000 frames per second of a laser powder bed fusion processes for stainless steel. The laser power and speed parameters are systematically varied. The goal is to reconstruct the power and speed of the laser from this video input. By comparing the predicted laser parameters with the pre-set parameters of the machine producing the laser, anomalies in the printing process can be detected faster, leading to more efficient quality control. The base \ac{ML} model used for this task is a SwinTransformer~\cite{yang2023swin3dpretrainedtransformerbackbone}, with the \ac{HPO} search space consisting of multiple, Transformer-specific architectural and optimizer-related parameters, such as the number of attention heads, see Tab.~\ref{tab:search_space}. The model is trained on the C027 cylinder with a 80/20 split for training and validation and is approximately 60 \ac{GB} in size. It is evaluated on the C028 cylinder for testing purposes. The \ac{MSE} between predicted and actual laser power and speed is computed to assess the accuracy of the SwinTransformer.
All models are trained for $\min\_t = 5$ to $\max\_t$ $= 20$ epochs and a reduction and scaling factor $\mathtt{sf}=\mathtt{rf}=2$ is chosen for the schedulers. Following Eq.~\eqref{eq:rung_milestones}, this results in rung milestones at epochs $5, 10,$ and $20$. 

\subsection{Computational Fluid Dynamics}
The \ac{CFD} dataset contains actuated turbulent boundary layer flow data, generated from a simulation~\cite{Albers:996125}. 
The \ac{CFD} dataset is stored in HDF5 file format and comprises several widths. In this study, widths of $1{,}000, 1{,}200,$ and $1{,}600$ are used as training dataset (approximately $4.8$ \ac{TB} in size). Width of $1{,}800$ and $3{,}000$ are used as validation and test datasets (approximately $3.5$ \ac{TB} in size) to assess extrapolation performance. Altogether, the dataset is approximately $8.3$ \ac{TB} in size.

A convolutional autoencoder, selected from the AI4HPC repository~\cite{ai4hpc, inanc2023library}, is employed for flow reconstruction. The autoencoder comprises an encoder, a decoder, and a latent space representing a compressed, lower-dimensional version of the input. Both the encoder and decoder include four convolutional layers. In the encoder, the initial two layers perform down-sampling to compress the data, while in the decoder, they perform up-sampling to decompress the data in the latent space. The remaining layers perform regular convolution. 
The \ac{HPO} search space consists of the type of activation function as an architectural parameter and several optimizer-related ones, see Tab.~\ref{tab:search_space}. 

The autoencoders are trained for $\min\_t = 5$ to $\max\_t = 40$ epochs and a reduction and scaling factor $\mathtt{sf}=\mathtt{rf}=2$ is chosen for the schedulers. Following Eq.~\eqref{eq:rung_milestones}, this results in rung milestones at epochs $5, 10, 20,$ and $40$. 

The \ac{MSE} between the input and the reconstructed output flow field is computed and used to assess the accuracy of the autoencoders.  As a further measure of solution quality, also the relative reconstruction error is computed on the test set.

\quad

All experiments use reduction and scaling factors $\mathtt{sf}=\mathtt{rf}=2$, as this provides a suitable trade-off between terminating unpromising trials and scaling up \ac{GPU} resources. Choosing a scaling factor that is too large can lead to learning instabilities at the rung milestones, due to the abrupt increase in the number of \acp{GPU} allocated. Similarly, selecting a reduction factor that is too large may result in prematurely terminating trials that could have shown strong performance in later training stages. The value $\min_t = 5$ is selected to ensure sufficient initial training and to avoid interference with the learning rate warm-up phase, which can last up to five epochs (see Tab.~\ref{tab:search_space}). Terminating trials during this period could result in inaccurate early stopping decisions. The values $\max_t = 20$ and $\max_t = 40$ are chosen to provide adequate training time for convergence, while also considering computational resource constraints.

\section{Results}\label{sec:results}
This section presents the experimental results of running the proposed algorithm on two supercomputer systems, which are introduced in Sec.~\ref{ss:hpc_systems}. Section~\ref{sec:scaling} focuses on the scaling performance of the \ac{RASDA} algorithm on up to 1{,}024 \acp{GPU}, while Sec.~\ref{sec:speedups_accuracy} compares the \ac{RASDA} against the plain \ac{ASHA} scheduler without any resource adaptation. Section~\ref{sec:large_scale} reports the performance of \ac{RASDA} at the large scale, and in Sec.~\ref{sec:ablation_studies} different ablation experiments are presented.

\subsection{Supercomputers}
\label{ss:hpc_systems}
The two supercomputer modules used for the experiments in this study are both located at the Jülich Supercomputing Centre. 

The first system is the JURECA-DC-GPU module~\cite{juelich2021jureca} consisting of a total of 192 accelerated compute nodes. Each node is equipped with two AMD EPYC 7742 \acsp{CPU} with 128 cores clocked at $2.25$~GHz and four NVIDIA A100 \acsp{GPU}, each with 40~GB high-bandwidth memory. 
The second \ac{HPC} system is the JUWELS BOOSTER module~\cite{juwels_booster} consisting of a total of 936 compute nodes. Each node is equipped with two AMD EPYC Rome 7402 \acsp{CPU} with 48 cores clocked at $2.8$~GHz, and four NVIDIA A100 \ac{GPU} with $40$~GB high-bandwidth memory. 
The main difference between the two systems is the number of InfiniBand interconnects: the JURECA-DC-GPU system features only two per node, while the JUWELS BOOSTER has four per node and therefore a higher network transmission bandwidth. 

As of June 2024, both supercomputers are among the top $10\%$ most energy-efficient supercomputers in the world, according to the GREEN500 list\footnote{GREEN500:
\tiny{\url{https://top500.org/lists/green500/list/2024/06/}}}.  

\subsection{Scaling Performance}\label{sec:scaling}
To evaluate the scalability of the \ac{RASDA} algorithm, two weak scaling experiments, where the number of \ac{HPO} configurations to evaluate is increased with the number of \acp{GPU}, are conducted with a lower number of training epochs. For this purpose, the \ac{CV} application case as a representative benchmark for \ac{DL} workloads is selected. It should be noted that while the asynchronous nature of the plain \ac{ASHA} algorithm naturally leads to good scalability~\cite{ASHA}, the goal of this study is to demonstrate that the additional resource allocation mechanism in \ac{RASDA} maintains this favorable scaling behavior.

\begin{figure*}
    \centering
    \includegraphics[width=0.49\textwidth]{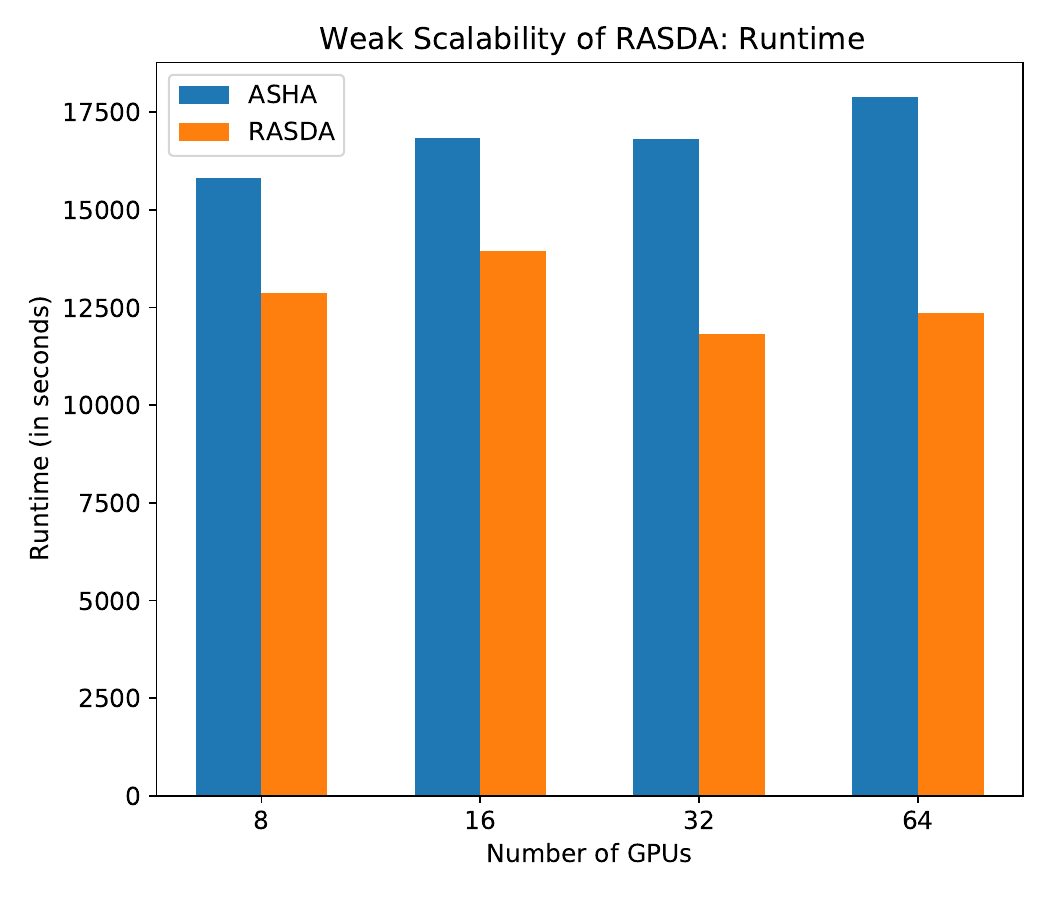}
    \includegraphics[width=0.49\textwidth]{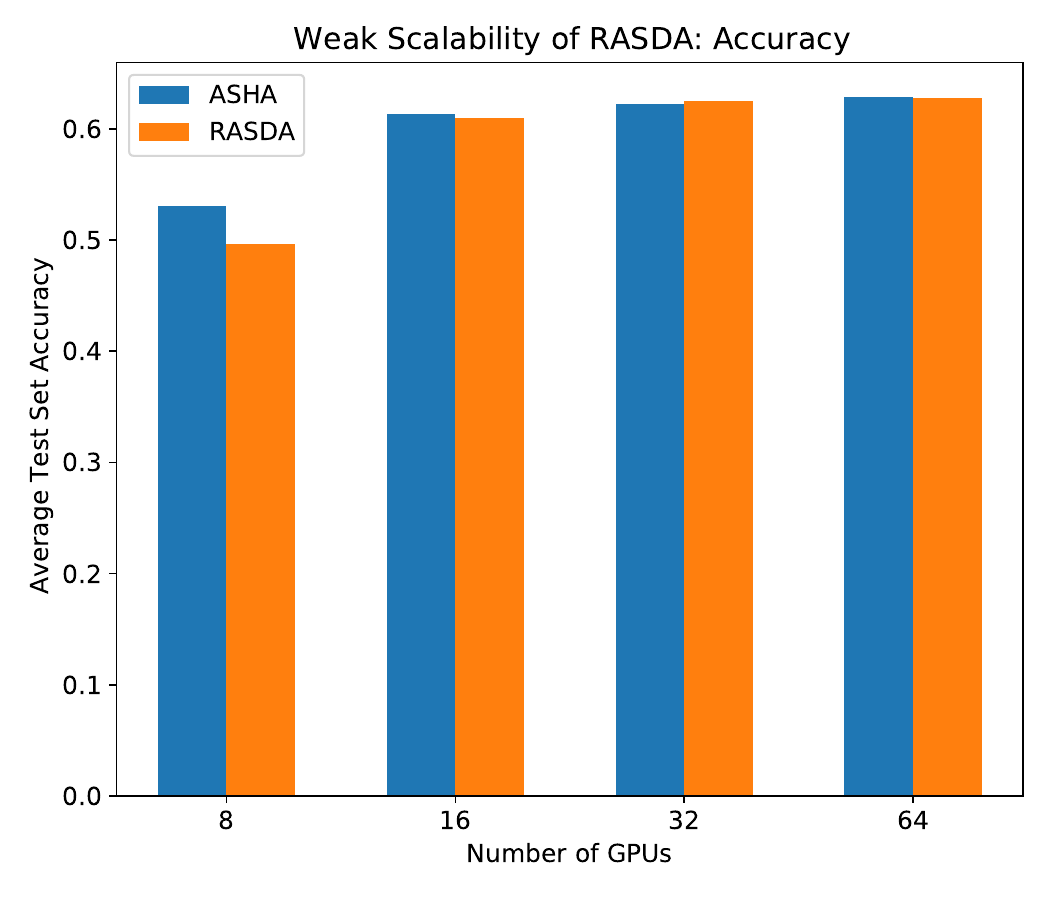}
    \caption{Comparison of \ac{ASHA} and \ac{RASDA} for training a ResNet50 model on ImageNet for 20 epochs on different scales on the JURECA-DC-GPU system.}
    \label{fig:plots_at_scale}
\end{figure*}

The first weak scaling experiment considers a smaller scale of $8$ to $64$ \acp{GPU}. The runtime and accuracy of the \ac{RASDA} algorithm is compared to the plain \ac{ASHA} algorithm for training a ResNet50 on the ImageNet dataset for 20 epochs, see Fig.~\ref{fig:plots_at_scale}.  It can be seen that on all scales (from 8 to 64 \acp{GPU}), the \ac{RASDA} algorithm achieves consistently lower runtimes up to a factor of $1.45$ faster than its \ac{ASHA} counterpart while matching the final test set accuracy in almost all cases. \textcolor{black}{Reduced test set accuracy is observed only at $8$ and, to a lesser extent, $16$ \acp{GPU}, as in these small settings the number of hyperparameter configurations evaluated is low and \ac{RASDA} cannot yet fully benefit from large batch size training (see Sec.~\ref{sec:large_bs_training}).} 

\begin{figure}
    \centering
    \includegraphics[width=0.49\textwidth]{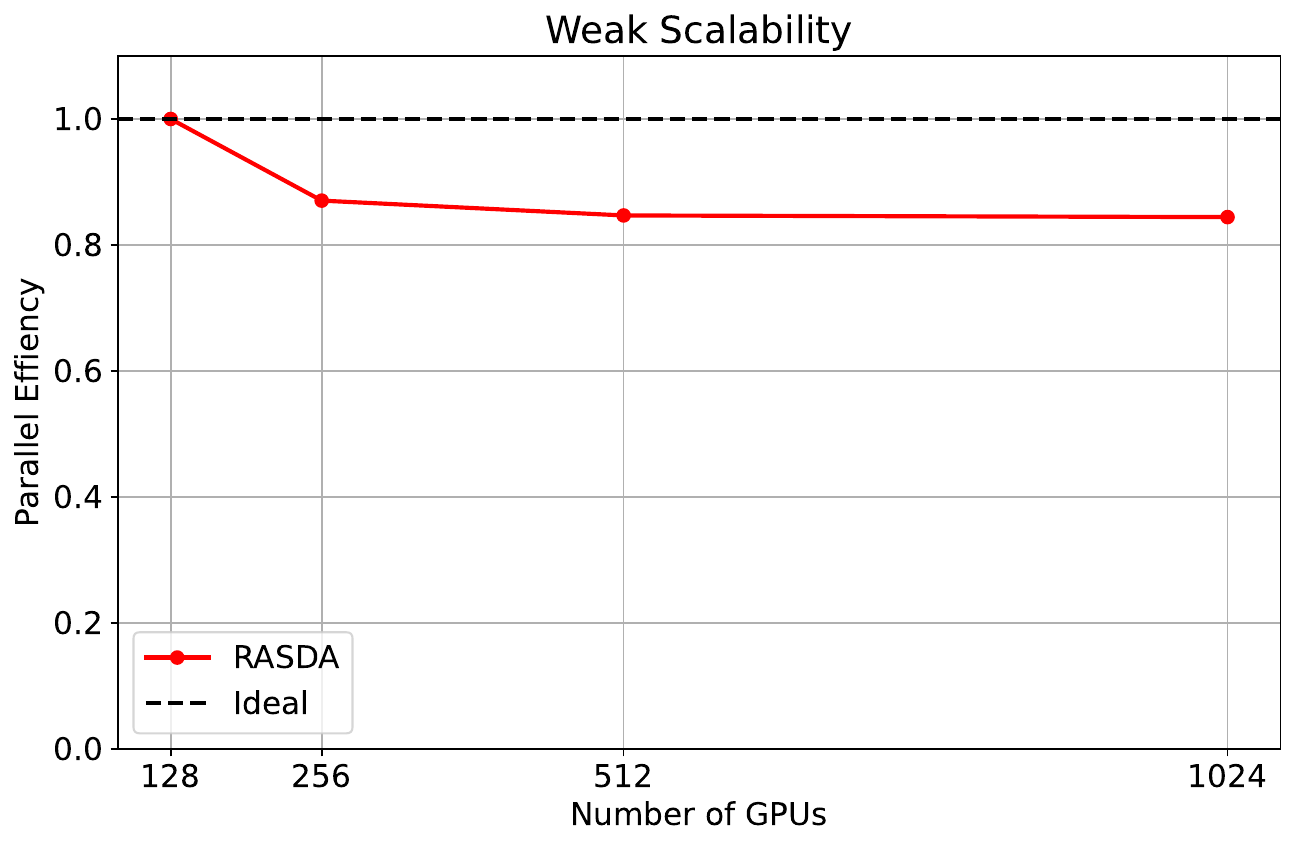}
    \caption{Weak scalability of the \ac{RASDA} algorithm on up to $1{,}024$ \acp{GPU} on the JUWELS BOOSTER system, including ideal scalability for comparison.}
    \label{fig:weak_scalability}
\end{figure}

The second scaling experiment considers a large scale of $128$ to $1{,}024$ \acp{GPU}, see Fig.~\ref{fig:weak_scalability}. The weak scalability of the \ac{RASDA} algorithm is evaluated by training a ResNet for six epochs. The results show that the algorithm maintains a high parallel efficiency of $>0.84$ on up to $1{,}024$ \acp{GPU}. 

It should be noted that strong scaling experiments that keep the number of hyperparameter configurations consistent across all scales are generally infeasible for this type of \ac{HPO} workload, as evaluating a large number of configurations on a small number of \acp{GPU} would take too long. 

The better scaling performance of \ac{RASDA} compared to \ac{ASHA} can also be observed when examining specific \ac{HPC} metrics. \ac{RASDA} consistently achieves higher \ac{GPU} utilization, as it avoids idling \acp{GPU}, see Fig.~\ref{fig:algorithm_comparison}. \ac{RASDA} achieves approximately $\approx 80\%$ \ac{GPU} utilization, compared to around $\approx 54\%$ for the \ac{ASHA} case. Since more \acp{GPU} are actively used, this also leads to higher I/O demand compared to \ac{ASHA}, where idle \acp{GPU} consume less data. However, because \ac{RASDA} achieves significantly shorter runtimes, it is expected to be more energy-efficient overall.

\subsection{Speed-Ups and Accuracy}\label{sec:speedups_accuracy}
\begin{figure*}[!h]
    \centering
    \includegraphics[width=0.32\textwidth]{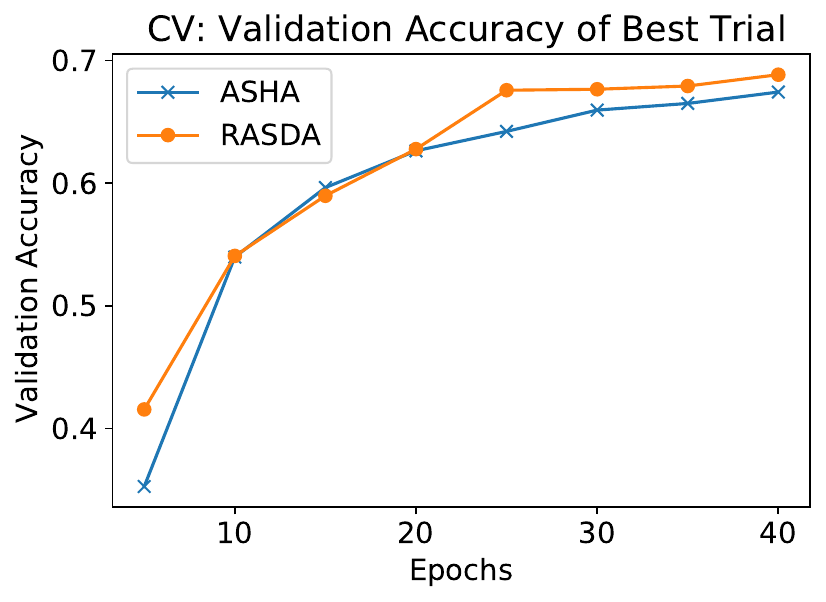}
    \includegraphics[width=0.35\textwidth]{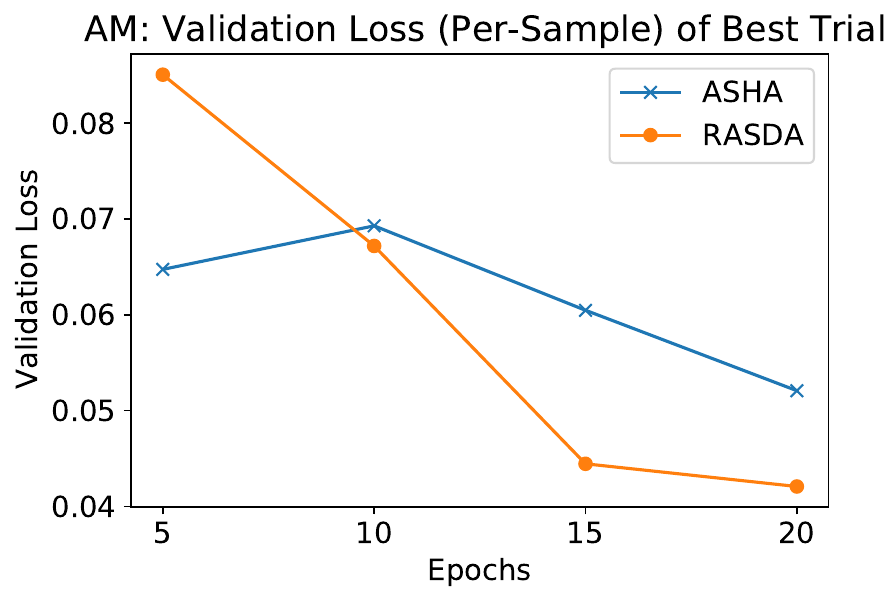}
    \includegraphics[width=0.32\textwidth]{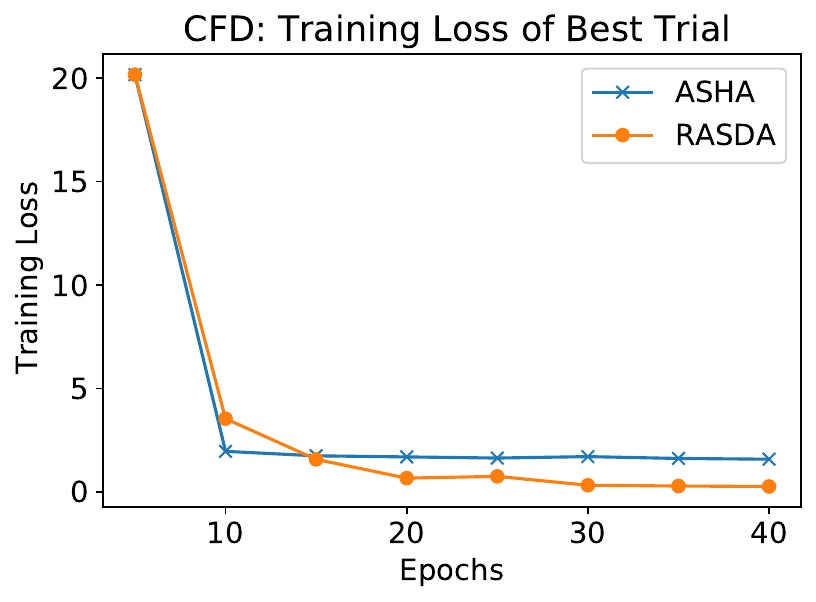}
    \caption{Exemplary comparison of the performance (in terms of validation accuracy, training loss, and validation loss) of the best configuration found by \ac{ASHA} and \ac{RASDA} for the different application cases.}
    \label{fig:solution_quality_over_epochs}
\end{figure*}

\begin{figure*}[!h]
    \centering
    \includegraphics[width=0.33\linewidth]{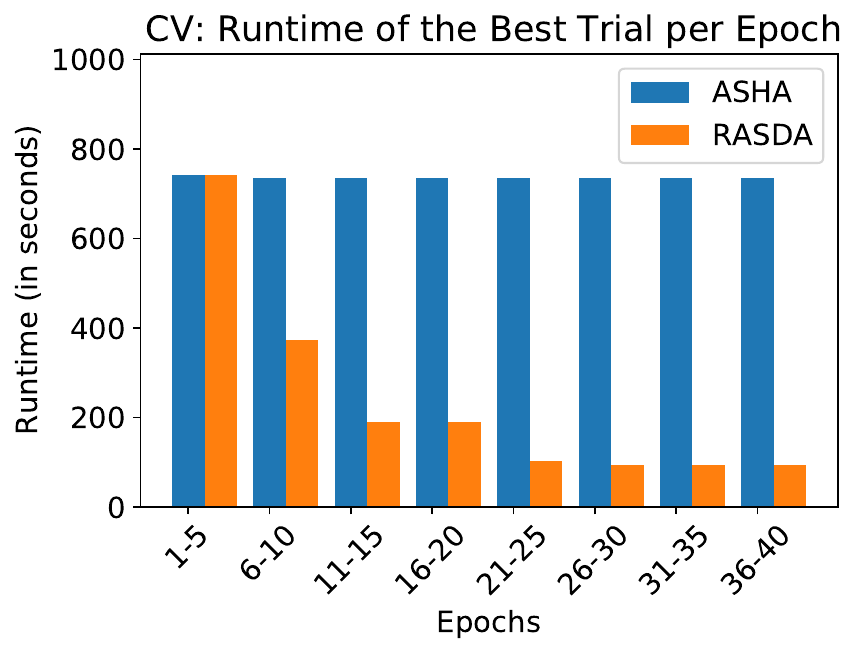}
    \includegraphics[width=0.33\linewidth]{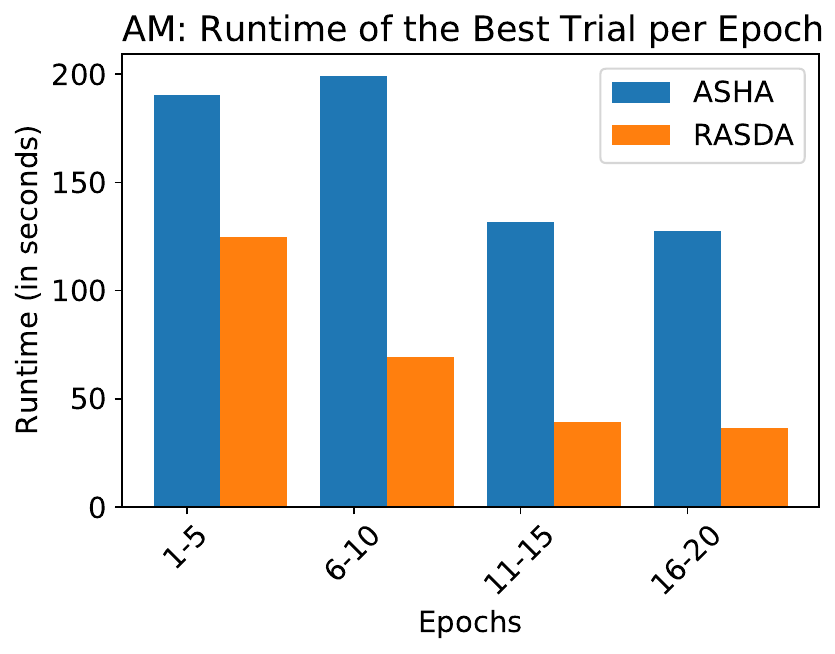}
    \includegraphics[width=0.33\linewidth]{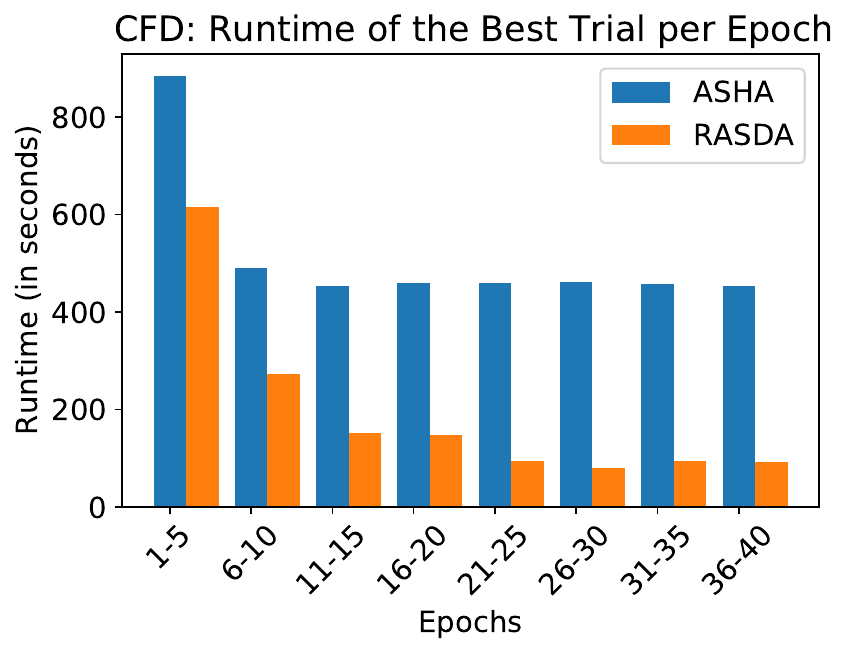}
    \caption{Exemplary comparison of the runtime per epoch of the best configuration found by \ac{ASHA} and \ac{RASDA} for the different application cases. Note that for the \ac{CFD} and \ac{AM} case, the architectural parameters chosen by the respective \ac{HPO} method also influence the model size, which is why here differences in runtime can be observed already during the first five epochs.}
    \label{fig:runtime_per_epoch}
\end{figure*}

\begin{figure*}[!h]
    \centering
    \includegraphics[width=0.33\linewidth]{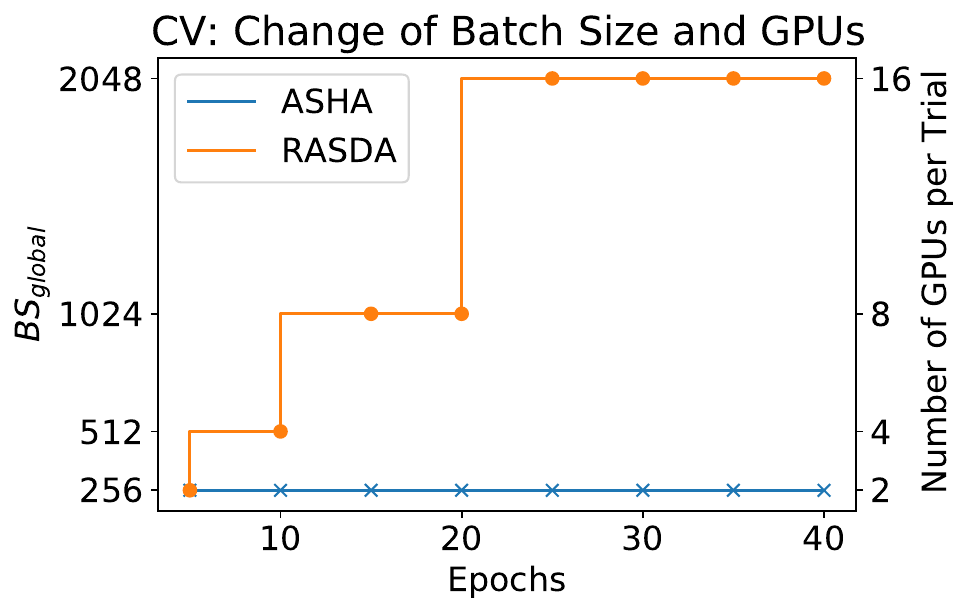}
    \includegraphics[width=0.33\linewidth]{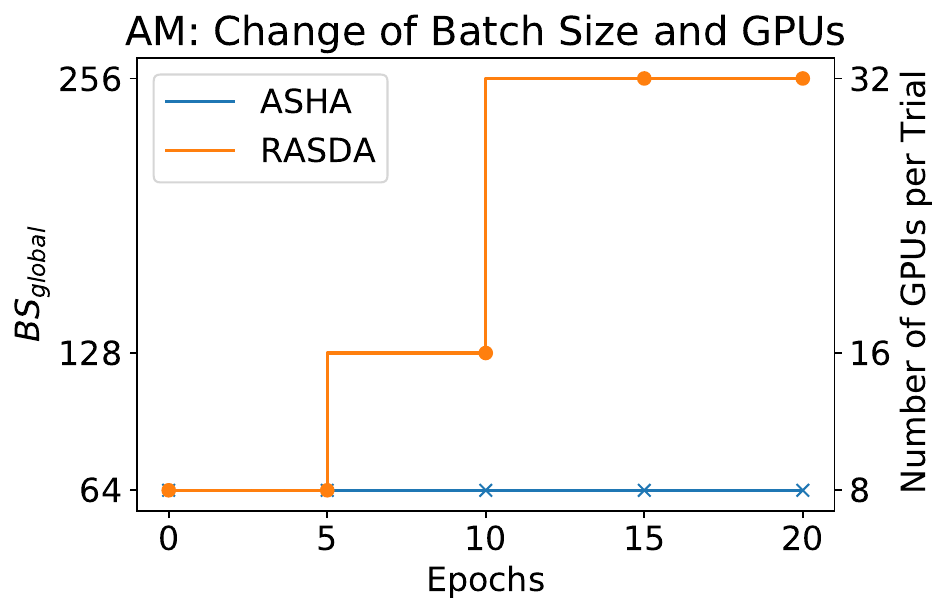}
    \includegraphics[width=0.33\linewidth]{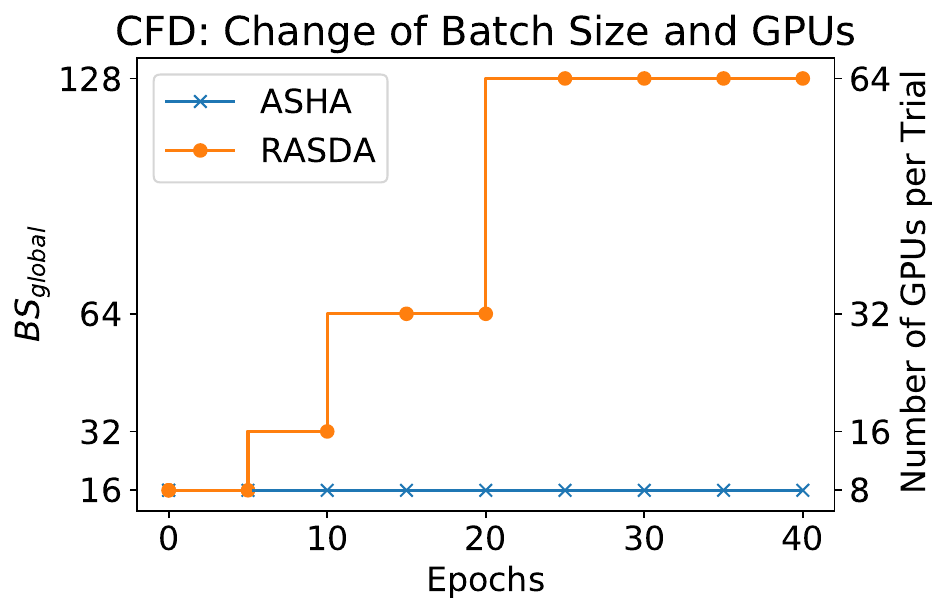}
    \caption{Comparison of the global batch size and the number of \acp{GPU} per trial for \ac{ASHA} and \ac{RASDA} for the different application cases.}
    \label{fig:bs_per_epoch}
\end{figure*}

To evaluate the performance of the \ac{RASDA} algorithm in terms of speed-up and accuracy and to juxtapose it to the plain \ac{ASHA} algorithm considering the application cases, the number of training epochs is increased within the $\min\_t$ and $\max\_t$ range specified in Sec.~\ref{sec:application_cases}. The general results for the three application case, averaged over three different runs for all application cases, are presented in Tab.~\ref{tab:cv_results}, Tab.~\ref{tab:am_results}, and Tab.~\ref{tab:cfd_results}. The solution quality over time is presented in Fig.~\ref{fig:solution_quality_over_epochs},
an in-depth performance analysis of the runtimes per epoch is given in  Fig.~\ref{fig:runtime_per_epoch}, and the change of batch size and number of \acp{GPU} per trial is depicted in Fig.~\ref{fig:bs_per_epoch}. The results correspond to an exemplary best-performing trial from one of the three runs. The following paragraphs provide a more detailed discussion of these tables and figures.

% CV case
For the \ac{CV} application case, a total of 32 hyperparameter combinations are evaluated simultaneously on 64 \acp{GPU} on the JURECA-DC-GPU system, with each parallel trial starting with two \acp{GPU}. Compared to the plain \ac{ASHA} approach, the \ac{RASDA} algorithm reduces the overall average runtime of the \ac{HPO} process by a factor of $\approx 1.71$ from $527$ to $308$ minutes, see Tab.~\ref{tab:cv_results}. The average solution quality, i.e., the training, validation, and test set accuracy of the best trial discovered during the process, slightly outperforms the ones of the plain \ac{ASHA}. This indicates that scaling the batch size and the learning rate during the training process does not impact the learning process in this case. A closer look at one of the best-performing trials in Fig.~\ref{fig:runtime_per_epoch} reveals that indeed the average runtime decreases in the \ac{RASDA} case once the resource adaptation in space sets in after the first five epochs. As can be seen in Fig.~\ref{fig:bs_per_epoch}, $BS_{global}$ increases from 256 to 2{,}048 during the training and the number of \acp{GPU} from 2 to 16 per trial for the \ac{RASDA} case, while both stay constant in the plain \ac{ASHA} case. The plot of the validation accuracy over the number of epochs in Fig.~\ref{fig:solution_quality_over_epochs} confirms that \ac{RASDA} slightly outperforms the \ac{ASHA} approach in terms of solution quality.

% AM case
For the \ac{AM} application case, the \ac{HPO} process evaluates 16 configurations, using a total of 128 \acp{GPU} on the JURECA-DC-GPU system. The trials start out with 8 \acp{GPU} each, which increases to 32 \acp{GPU} for the top-performing trials, at the same time increasing $BS_{global}$ from 64 to 256. As the models are only trained for a total amount of 20 epochs (due to the long training times of transformer models), only two resource-doubling steps, i.e., at epoch 5 and epoch 10, take place, see Fig.~\ref{fig:bs_per_epoch}. Table~\ref{tab:am_results} provides an overview of the results in terms of runtime and solution quality. In comparison with the plain \ac{ASHA} algorithm, a speed-up by a factor of $1.52$ is achieved, reducing the required \ac{HPO} runtime of the models from 96 to 63 minutes. On both the validation and test dataset, the best configuration found by \ac{RASDA} again outperforms the one found with the plain \ac{ASHA} after 20 epochs, as can be seen in Fig.~\ref{fig:solution_quality_over_epochs}.

% CFD case
The \ac{CFD} application case features the largest dataset used in this study. The whole \ac{HPO} process evaluates 16 configurations on 128 \acp{GPU} simultaneously on the JUWELS BOOSTER module. Each trial starts with 8 \acp{GPU}, which is increased over time to 64 \ac{GPU} by the \ac{RASDA} algorithm. As can be seen from Tab.~\ref{tab:cfd_results}, the most significant speed-up with a factor of $\approx 1.9$ is achieved in this case, with \ac{RASDA} reducing the runtime of the \ac{HPO} process from 325 to 170 minutes. In this case, also the average \ac{MSE} decreases by a factor of $\approx 1.88$. This is likely due to the even better generalization capabilities caused by increasing the batch size over time (following the insights explained in Sec.~\ref{sec:large_bs_training}). Obviously, this outperforms just annealing of the learning rate. This observation is in line with the findings of Smith et~al.~\cite{smith2018dont}. \ac{RASDA} also achieves a low relative reconstruction error of just $1.15 \%$ on the test set.

In general, the most substantial speed-up is established on the largest dataset from the \ac{CFD} domain. This is expected, as with a larger dataset, the benefit of adding more \acp{GPU} to the data-parallel training loop also increases. 
It is additionally interesting to observe that the speed-ups can be attained on both the JURECA-DC-GPU and JUWELS BOOSTER systems, although the latter features twice the network bandwidth. While \ac{RASDA} already yields substantial benefits on JURECA-DC-GPU with its moderate network infrastructure, the doubled network bandwidth of JUWELS BOOSTER further amplifies these speed-ups, highlighting how the approach particularly profits from fast interconnects.


\begin{table}[h]
\caption{\ac{HPO} for the \ac{CV} application case, trained for 40 epochs on 64 \acp{GPU} on the JURECA-DC-GPU system. Results are averaged over three random seeds. Better results ($\uparrow$ or $\downarrow$ depending on the metric) are underlined.}
\begin{tabular}{l|c|c|c}
\textbf{Metric} & \textbf{ASHA} & \textbf{RASDA} & \textbf{Diff.} \\
\hline
Train Accuracy $\uparrow$         & 0.6976            & \underline{0.7310}              & $1.05 \times$                    \\ 
Val Accuracy $\uparrow$        & 0.6728             & \underline{0.6813}                & $1.01 \times$                        \\
Test Accuracy $\uparrow$       & 0.6688              & \underline{0.6766}                & $1.01 \times$                       \\
Runtime (in seconds) $\downarrow$    & 31637            & \underline{18502}              & $1.71 \times$                        
\end{tabular}
\label{tab:cv_results}
\end{table}

\begin{table}[h]
\caption{\ac{HPO} for the \ac{AM} application case, trained for 20 epochs on 128 \acp{GPU} on the JURECA-DC-GPU system. Results are averaged over three random seeds. Better results ($\uparrow$ or $\downarrow$ depending on the metric) are underlined. For better comparison the metrics were recomputed on a per-sample basis after the run.}
\begin{tabular}{l|c|c|c}
\textbf{Metric} & \textbf{ASHA} & \textbf{RASDA} & \textbf{Diff.} \\
\hline
Val MSE $\downarrow$        & 0.0455           & \underline{0.0404}                & $1.12 \times$                          \\
Test MSE $\downarrow$       & 0.0554              & \underline{0.0516}                & $1.07 \times$                        \\
Runtime (in seconds) $\downarrow$    & 5784            & \underline{3803}     &   $1.52 \times$                     
\end{tabular}
\label{tab:am_results}
\end{table}

\begin{table}[h]
\caption{\ac{HPO} for the \ac{CFD} application case, trained for 40 epochs on 128 \acp{GPU} on the JUWELS BOOSTER module. Results are averaged over three random seeds. Better results ($\uparrow$ or $\downarrow$ depending on the metric) are underlined.}
\begin{tabular}{l|c|c|c}
\textbf{Metric} & \textbf{ASHA} & \textbf{RASDA} & \textbf{Diff.} \\
\hline
Val MSE $\downarrow$ & \num{5.28e-06}             & \underline{\num{2.81e-06}}               & $1.88 \times$                        \\
Test MSE $\downarrow$ & \num{4.42e-06}              & \underline{\num{2.40E-06}}               & $1.84 \times$                         \\
Test Relative Error $\downarrow$ & 0.0185            & \underline{0.0115}                & $1.61 \times$                        \\
Runtime (in seconds) $\downarrow$       & 19487               & \underline{10242}              & $1.90 \times$                        
\end{tabular}
\label{tab:cfd_results}
\end{table}

\subsection{Performance at 1{,}024 \acp{GPU} Scale}\label{sec:large_scale}
While the superiority of \ac{RASDA} over plain \ac{ASHA} has been confirmed in the previous experiments using 64 and 128 \acp{GPU}, a final \ac{RASDA} experiment on a 1{,}024 \ac{GPU} scale is conducted on the JUWELS BOOSTER system. Again, using the \ac{CFD} application case, the number of configurations to be evaluated is increased to 64, with each trial starting with 16 \acp{GPU}. The models are trained for $\min\_t = 5$ and $\max\_t = 20$ epochs. The \ac{HPO} run took three hours and resulted in an improved model with a validation \ac{MSE} of $ \approx \num{3.63e-07}$, a test \ac{MSE} of $\approx \num{4.88e-08}$ and a relative test error of $\approx 0.0016$. Depending on the metric, this is a $7$ to $49$ times increase in solution quality, compared to the results of the \ac{HPO} run on 128 \acp{GPU} (see Tab.~\ref{tab:cfd_results_large_scale}), which highlights the potential of large-scale \ac{HPO} for scientific \ac{ML}.

\begin{table}[h]
\caption{Large-scale \ac{HPO} for the \ac{CFD} application case, evaluating $64$ configurations, trained for a maximum of 20 epochs on 1{,}024 \acp{GPU} on the JUWELS BOOSTER module, including relative improvement to the \ac{HPO} run on 128 \acp{GPU}.}
\begin{tabular}{l|c|c}
\textbf{Metric} & \textbf{RASDA - 1{,}024 \acp{GPU}} & \textbf{vs. 128 \acp{GPU}} \\
\hline
Val MSE  &  \num{3.63e-07}              & $7.74 \times$                        \\
Test MSE  &  \num{4.88e-08}              & $49.22 \times$                         \\
Test Rel. Error  &  0.0016             & $7.17 \times$                                      
\end{tabular}
\label{tab:cfd_results_large_scale}
\end{table}

\subsection{Ablation Studies}\label{sec:ablation_studies}
To evaluate the impact of different parameters on the performance of the \ac{RASDA} method, two ablation studies are conducted.
\subsubsection{Impact of Reduction and Scaling Factors}
All prior experiments use a scaling and reduction factor of $\mathtt{sf} = \mathtt{rf} = 2$. In this ablation study on the \ac{CV} application case, these factors are increased to $\mathtt{sf} = \mathtt{rf} = 4$. Since the number of \ac{GPU} accelerators per node on \ac{HPC} systems typically follows a power-of-two configuration, allocating partial nodes may lead to performance degradation. To ensure sufficient training time between decision points, the same values of $\min\_t = 5$ and $\max\_t = 20$ epochs are retained. According to Eq.~\eqref{eq:rung_milestones}, this results in two rung milestones at epochs $5$ and $20$.
As shown in Tab.~\ref{tab:reduction_factor_ablation}, \ac{RASDA} continues to outperform \ac{ASHA} under these settings, achieving higher accuracy and a runtime speed-up of $\approx 1.6\times$. However, this speed-up is smaller than the improvement observed with $\mathtt{sf} = \mathtt{rf} = 2$ (see Tab.~\ref{tab:cv_results}). While a larger scaling factor enables more aggressive allocation of \ac{GPU} resources to promising trials, it also increases the time between rung milestones due to the geometric progression (see Eq.~\ref{eq:rung_milestones}). This observation indicates that $\mathtt{rf} = 2$ is a suitable choice for the \ac{RASDA} scheduler.

\begin{table}
\caption{\textcolor{black}{Comparison of \ac{ASHA} and \ac{RASDA} with $\mathtt{sf}=\mathtt{rf}=4$ for the \ac{CV} application case, trained for 20 epochs on 64 \acp{GPU} on the JURECA-DC-GPU system. Results are averaged over three random seeds. Better results ($\uparrow$ or $\downarrow$ depending on the metric) are underlined.}}
\begin{tabular}{l|c|c|c}
\textcolor{black}{\textbf{Metric}} & \textcolor{black}{\textbf{\ac{ASHA}}} & \textcolor{black}{\textbf{\ac{RASDA}}} & \textcolor{black}{\textbf{Diff.}} \\
\hline
\textcolor{black}{Train Accuracy $\uparrow$} & \textcolor{black}{0.6179} & \textcolor{black}{\underline{0.6556}} & \textcolor{black}{$1.06 \times$} \\ 
\textcolor{black}{Val Accuracy $\uparrow$} & \textcolor{black}{0.6312} & \textcolor{black}{\underline{0.6348}} & \textcolor{black}{$1.01 \times$} \\
\textcolor{black}{Test Accuracy $\uparrow$} & \textcolor{black}{0.6250} & \textcolor{black}{\underline{0.6340}} & \textcolor{black}{$1.01 \times$} \\
\textcolor{black}{Runtime (in seconds) $\downarrow$} & \textcolor{black}{17830} & \textcolor{black}{\underline{11157}} & \textcolor{black}{$1.60 \times$}
\end{tabular}
\label{tab:reduction_factor_ablation}
\end{table}

\begin{table}
\caption{\textcolor{black}{Comparison of \ac{RASDA} and \ac{BOHB} on the \ac{CV} application case, trained for 20 epochs on 64 \acp{GPU} on the JURECA-DC-GPU system, evaluating $32$ hyperparameter samples. Results are averaged over three random seeds. Better results ($\uparrow$ or $\downarrow$ depending on the metric) are underlined.}}
\begin{tabular}{l|c|c|c}
\textcolor{black}{\textbf{Metric}} & \textcolor{black}{\textbf{BOHB}} & \textcolor{black}{\textbf{RASDA}} & \textcolor{black}{\textbf{Diff.}} \\
\hline
\textcolor{black}{Train Accuracy $\uparrow$} & \textcolor{black}{0.6130} & \textcolor{black}{\underline{0.6480}} & \textcolor{black}{$1.06 \times$} \\ 
\textcolor{black}{Val Accuracy $\uparrow$} & \textcolor{black}{0.6253} & \textcolor{black}{\underline{0.6271}} & \textcolor{black}{$1.00 \times$} \\
\textcolor{black}{Test Accuracy $\uparrow$} & \textcolor{black}{0.6222} & \textcolor{black}{\underline{0.6254}} & \textcolor{black}{$1.01 \times$} \\
\textcolor{black}{Runtime (in seconds) $\downarrow$} & \textcolor{black}{16825} & \textcolor{black}{\underline{11815}} & \textcolor{black}{$1.42 \times$}
\end{tabular}
\label{tab:bo_ablation}
\end{table}

\subsubsection{Comparison to \acf{BO}}
Although the primary comparison is between \ac{RASDA} and its closest scheduling-based counterpart, \ac{ASHA}, other types of \ac{HPO} and \ac{NAS} tools are also relevant. One commonly used method is \acf{BO}. A comparison between \ac{RASDA} and the \ac{BOHB} algorithm on the \ac{CV} application case is provided below. To function effectively, \ac{BO} requires the ability to generate new hyperparameter configurations based on past evaluations. As shown in Tab.~\ref{tab:bo_ablation}, when both methods are evaluated with the same number of hyperparameter samples, \ac{RASDA} outperforms \ac{BOHB} in terms of accuracy and runtime. These results suggest that \ac{RASDA} achieves a more favorable trade-off between runtime and solution quality compared to traditional \ac{BO}-based approaches.

\section{Summary and Outlook}\label{sec:summary}
\acs{RASDA}, a novel resource-adaptive successive doubling algorithm for \ac{HPO}, suitable for running on \ac{HPC} systems, was introduced. The key idea is to not only perform successive halving in time and let promising configurations train for longer (as is already the case in plain \ac{ASHA}), but to combine it with successive doubling in space and allocate more computational resources to the data-parallel training of promising configurations. 

The \ac{RASDA} method was evaluated extensively on a standard benchmarking task in the \ac{CV} domain as well as on two large datasets (up to 8.3 \ac{TB} in size) from the \ac{CFD} and \ac{AM} domains. The results confirm that \ac{RASDA} leads in these cases to speed-ups up to a factor of $\approx 1.9$ in comparison to the \ac{ASHA} algorithm. 

Another property of \ac{RASDA} is that it progressively scales up the global batch size of the trials as it adds more \acp{GPU} to their training loops. This helps them to avoid the degradation in solution quality, which is usually associated with large batch training. Remarkably, the approach did enhance the solution quality, aligning with literature findings suggesting that increasing the batch size can match or surpass the effects of learning rate annealing.

In addition, this study represents the first application of systematic \ac{HPO} to a scientific dataset at the \ac{TB} scale. A comparison of the application of \ac{RASDA} on 128 and 1024 \acp{GPU} revealed a significant improvement in model performance. Specifically, the larger-scale application identifies a model that is significantly superior in solution quality. These results demonstrate the scalability and efficiency of the \ac{RASDA} method, thus paving the way for the application of \ac{HPO} methods on current and future Exascale supercomputers. \textcolor{black}{It should be noted, however, that \ac{RASDA} relies heavily on efficient distributed training. As such, its full benefits are realized primarily on \ac{HPC} systems equipped with accelerators such as \acp{GPU}. On non-accelerated systems or in settings with very limited communication bandwidth, potential inefficiencies may limit performance. }

For future work, the optimal timing for scaling the batch size (through the addition of more \acp{GPU} to the data-parallel training loop) should be investigated more thoroughly. Furthermore, the impact of scaling the batch size on various hyperparameters beyond the learning rate (such as the weight decay values) warrants a deeper exploration.

\section*{Acknowledgements}
The authors thank Kurt De Grave and Cyril Blanc for contributing an implementation of the \ac{AM} use case.

This research has been performed in the CoE RAISE project, which received funding from the European Union’s \textit{Horizon 2020 – Research and Innovation Framework Programme} H2020-INFRAEDI-2019-1 under grant agreement no.~951733.

The authors gratefully acknowledge computing time on the supercomputer JURECA~\cite{juelich2021jureca} at Forschungszentrum Jülich under grant no. raise-ctp2. The authors gratefully acknowledge the Gauss Centre for Supercomputing e.V.\ (\href{https://www.gauss-centre.eu/}{www.gauss-centre.eu}) for funding this project by providing computing time through the John von Neumann Institute for Computing (NIC) on the GCS Supercomputer JUWELS~\cite{JUWELS} at Jülich Supercomputing Centre (JSC).

\section*{Conflict of Interest}
The authors declare that there are no conflicts of interest related to this work.

\section*{Author Contributions}
MA: Conceptualization, investigation, methodology, software, visualization, validation, writing - original draft. RS: Methodology, software, validation, writing - review and editing. HN: Supervision, validation, writing - review and editing. MR: Supervision, validation, writing - review and editing. AL: Supervision, validation, writing - review and editing, funding acquisition. 
%\newpage
\bibliographystyle{elsarticle-harv} 
\bibliography{bibliography}

\begin{thebibliography}{49}
\expandafter\ifx\csname natexlab\endcsname\relax\def\natexlab#1{#1}\fi
\providecommand{\url}[1]{\texttt{#1}}
\providecommand{\href}[2]{#2}
\providecommand{\path}[1]{#1}
\providecommand{\DOIprefix}{doi:}
\providecommand{\ArXivprefix}{arXiv:}
\providecommand{\URLprefix}{URL: }
\providecommand{\Pubmedprefix}{pmid:}
\providecommand{\doi}[1]{\href{http://dx.doi.org/#1}{\path{#1}}}
\providecommand{\Pubmed}[1]{\href{pmid:#1}{\path{#1}}}
\providecommand{\bibinfo}[2]{#2}
\ifx\xfnm\relax \def\xfnm[#1]{\unskip,\space#1}\fi
%Type = Inproceedings
\bibitem[{Aach et~al.(2023)Aach, Sarma, Inanc, Riedel and
  Lintermann}]{mixed_precision_hpo}
\bibinfo{author}{Aach, M.}, \bibinfo{author}{Sarma, R.},
  \bibinfo{author}{Inanc, E.}, \bibinfo{author}{Riedel, M.},
  \bibinfo{author}{Lintermann, A.}, \bibinfo{year}{2023}.
\newblock \bibinfo{title}{Short paper: Accelerating hyperparameter optimization
  algorithms with mixed precision}, in: \bibinfo{booktitle}{Proceedings of the
  SC '23 Workshops of The International Conference on High Performance
  Computing, Network, Storage, and Analysis}, \bibinfo{publisher}{ACM}. p.
  \bibinfo{pages}{1776–1779}.
\newblock \DOIprefix\doi{10.1145/3624062.3624259}.
%Type = Inproceedings
\bibitem[{Aach et~al.(2022)Aach, Sedona, Lintermann, Cavallaro, Neukirchen and
  Riedel}]{aach_large_bs}
\bibinfo{author}{Aach, M.}, \bibinfo{author}{Sedona, R.},
  \bibinfo{author}{Lintermann, A.}, \bibinfo{author}{Cavallaro, G.},
  \bibinfo{author}{Neukirchen, H.}, \bibinfo{author}{Riedel, M.},
  \bibinfo{year}{2022}.
\newblock \bibinfo{title}{Accelerating hyperparameter tuning of a deep learning
  model for remote sensing image classification}, in:
  \bibinfo{booktitle}{IGARSS 2022 - 2022 IEEE International Geoscience and
  Remote Sensing Symposium}, \bibinfo{publisher}{IEEE}. pp.
  \bibinfo{pages}{263--266}.
\newblock \DOIprefix\doi{10.1109/IGARSS46834.2022.9883257}.
%Type = Inproceedings
\bibitem[{Akiba et~al.(2019)Akiba, Sano, Yanase, Ohta and Koyama}]{optuna}
\bibinfo{author}{Akiba, T.}, \bibinfo{author}{Sano, S.},
  \bibinfo{author}{Yanase, T.}, \bibinfo{author}{Ohta, T.},
  \bibinfo{author}{Koyama, M.}, \bibinfo{year}{2019}.
\newblock \bibinfo{title}{Optuna: A next-generation hyperparameter optimization
  framework}, in: \bibinfo{booktitle}{Proceedings of the 25th ACM SIGKDD
  International Conference on Knowledge Discovery \& Data Mining},
  \bibinfo{publisher}{ACM}. p. \bibinfo{pages}{2623–2631}.
\newblock \DOIprefix\doi{10.1145/3292500.3330701}.
%Type = Misc
\bibitem[{Albers et~al.(2023)Albers, Meysonnat, Fernex, Semaan, Noack,
  Schröder and Lintermann}]{Albers:996125}
\bibinfo{author}{Albers, M.}, \bibinfo{author}{Meysonnat, P.S.},
  \bibinfo{author}{Fernex, D.}, \bibinfo{author}{Semaan, R.},
  \bibinfo{author}{Noack, B.R.}, \bibinfo{author}{Schröder, W.},
  \bibinfo{author}{Lintermann, A.}, \bibinfo{year}{2023}.
\newblock \bibinfo{title}{{A}ctuated {T}urbulent {B}oundary {L}ayer {F}lows
  {D}ataset}.
\newblock \DOIprefix\doi{10.34730/5dbc8e35f21241d0889906136cf28d26}.
%Type = Inproceedings
\bibitem[{Balaprakash et~al.(2019)Balaprakash, Egele, Salim, Wild, Vishwanath,
  Xia, Brettin and Stevens}]{cancer_rf}
\bibinfo{author}{Balaprakash, P.}, \bibinfo{author}{Egele, R.},
  \bibinfo{author}{Salim, M.}, \bibinfo{author}{Wild, S.},
  \bibinfo{author}{Vishwanath, V.}, \bibinfo{author}{Xia, F.},
  \bibinfo{author}{Brettin, T.}, \bibinfo{author}{Stevens, R.},
  \bibinfo{year}{2019}.
\newblock \bibinfo{title}{Scalable reinforcement-learning-based neural
  architecture search for cancer deep learning research}, in:
  \bibinfo{booktitle}{Proceedings of the International Conference for High
  Performance Computing, Networking, Storage and Analysis},
  \bibinfo{publisher}{ACM}.
\newblock \DOIprefix\doi{10.1145/3295500.3356202}.
%Type = Inproceedings
\bibitem[{Balaprakash et~al.(2018)Balaprakash, Salim, Uram, Vishwanath and
  Wild}]{deep_hyper}
\bibinfo{author}{Balaprakash, P.}, \bibinfo{author}{Salim, M.},
  \bibinfo{author}{Uram, T.D.}, \bibinfo{author}{Vishwanath, V.},
  \bibinfo{author}{Wild, S.M.}, \bibinfo{year}{2018}.
\newblock \bibinfo{title}{Deephyper: Asynchronous hyperparameter search for
  deep neural networks}, in: \bibinfo{booktitle}{2018 IEEE 25th International
  Conference on High Performance Computing}, \bibinfo{publisher}{IEEE}. pp.
  \bibinfo{pages}{42--51}.
\newblock \DOIprefix\doi{10.1109/HiPC.2018.00014}.
%Type = Inproceedings
\bibitem[{Bergstra et~al.(2011)Bergstra, Bardenet, Bengio and
  K\'{e}gl}]{NIPS2011_86e8f7ab}
\bibinfo{author}{Bergstra, J.}, \bibinfo{author}{Bardenet, R.},
  \bibinfo{author}{Bengio, Y.}, \bibinfo{author}{K\'{e}gl, B.},
  \bibinfo{year}{2011}.
\newblock \bibinfo{title}{Algorithms for hyper-parameter optimization}, in:
  \bibinfo{editor}{Shawe-Taylor, J.}, \bibinfo{editor}{Zemel, R.},
  \bibinfo{editor}{Bartlett, P.}, \bibinfo{editor}{Pereira, F.},
  \bibinfo{editor}{Weinberger, K.Q.} (Eds.), \bibinfo{booktitle}{Proceedings of
  the 24th International Conference on Neural Information Processing Systems},
  \bibinfo{publisher}{Curran Associates, Inc.}
%Type = Article
\bibitem[{Bergstra and Bengio(2012)}]{random_search}
\bibinfo{author}{Bergstra, J.}, \bibinfo{author}{Bengio, Y.},
  \bibinfo{year}{2012}.
\newblock \bibinfo{title}{Random search for hyper-parameter optimization}.
\newblock \bibinfo{journal}{Journal of Machine Learning Research}
  \bibinfo{volume}{13}, \bibinfo{pages}{281–305}.
\newblock \URLprefix \url{http://jmlr.org/papers/v13/bergstra12a.html}.
%Type = Article
\bibitem[{Blanc et~al.(2023)Blanc, Ahar and {De Grave}}]{BLANC2023100161}
\bibinfo{author}{Blanc, C.}, \bibinfo{author}{Ahar, A.}, \bibinfo{author}{{De
  Grave}, K.}, \bibinfo{year}{2023}.
\newblock \bibinfo{title}{Reference dataset and benchmark for reconstructing
  laser parameters from on-axis video in powder bed fusion of bulk stainless
  steel}.
\newblock \bibinfo{journal}{Additive Manufacturing Letters}
  \bibinfo{volume}{7}, \bibinfo{pages}{100161}.
\newblock \DOIprefix\doi{https://doi.org/10.1016/j.addlet.2023.100161}.
%Type = Article
\bibitem[{Cowen-Rivers et~al.(2022)Cowen-Rivers, Lyu, Tutunov, Wang, Grosnit,
  Griffiths, Maraval, Jianye, Wang, Peters and Bou-Ammar}]{hebo}
\bibinfo{author}{Cowen-Rivers, A.I.}, \bibinfo{author}{Lyu, W.},
  \bibinfo{author}{Tutunov, R.}, \bibinfo{author}{Wang, Z.},
  \bibinfo{author}{Grosnit, A.}, \bibinfo{author}{Griffiths, R.R.},
  \bibinfo{author}{Maraval, A.M.}, \bibinfo{author}{Jianye, H.},
  \bibinfo{author}{Wang, J.}, \bibinfo{author}{Peters, J.},
  \bibinfo{author}{Bou-Ammar, H.}, \bibinfo{year}{2022}.
\newblock \bibinfo{title}{Hebo: Pushing the limits of sample-efficient
  hyper-parameter optimisation}.
\newblock \bibinfo{journal}{J. Artif. Int. Res.} \bibinfo{volume}{74}.
\newblock \URLprefix \url{https://doi.org/10.1613/jair.1.13643},
  \DOIprefix\doi{10.1613/jair.1.13643}.
%Type = Inproceedings
\bibitem[{Dunlap et~al.(2021)Dunlap, Kandasamy, Misra, Liaw, Jordan, Stoica and
  Gonzalez}]{SEER}
\bibinfo{author}{Dunlap, L.}, \bibinfo{author}{Kandasamy, K.},
  \bibinfo{author}{Misra, U.}, \bibinfo{author}{Liaw, R.},
  \bibinfo{author}{Jordan, M.}, \bibinfo{author}{Stoica, I.},
  \bibinfo{author}{Gonzalez, J.E.}, \bibinfo{year}{2021}.
\newblock \bibinfo{title}{Elastic hyperparameter tuning on the cloud}, in:
  \bibinfo{booktitle}{Proceedings of the ACM Symposium on Cloud Computing},
  \bibinfo{publisher}{ACM}. p. \bibinfo{pages}{33–46}.
\newblock \URLprefix \url{https://doi.org/10.1145/3472883.3486989},
  \DOIprefix\doi{10.1145/3472883.3486989}.
%Type = Inproceedings
\bibitem[{Falkner et~al.(2018)Falkner, Klein and Hutter}]{falkner-icml-18}
\bibinfo{author}{Falkner, S.}, \bibinfo{author}{Klein, A.},
  \bibinfo{author}{Hutter, F.}, \bibinfo{year}{2018}.
\newblock \bibinfo{title}{{BOHB}: Robust and efficient hyperparameter
  optimization at scale}, in: \bibinfo{booktitle}{Proceedings of the 35th
  International Conference on Machine Learning}, \bibinfo{publisher}{{PMLR}}.
  pp. \bibinfo{pages}{1436--1445}.
\newblock \URLprefix \url{https://proceedings.mlr.press/v80/falkner18a.html}.
%Type = Inbook
\bibitem[{Feurer and Hutter(2019)}]{Feurer2019}
\bibinfo{author}{Feurer, M.}, \bibinfo{author}{Hutter, F.},
  \bibinfo{year}{2019}.
\newblock \bibinfo{title}{Hyperparameter Optimization}.
  \bibinfo{publisher}{Springer International Publishing},
  \bibinfo{address}{Cham}.
\newblock pp. \bibinfo{pages}{3--33}.
\newblock \DOIprefix\doi{10.1007/978-3-030-05318-5_1}.
%Type = Inproceedings
\bibitem[{Glorot and Bengio(2010)}]{pmlr-v9-glorot10a}
\bibinfo{author}{Glorot, X.}, \bibinfo{author}{Bengio, Y.},
  \bibinfo{year}{2010}.
\newblock \bibinfo{title}{Understanding the difficulty of training deep
  feedforward neural networks}, in: \bibinfo{editor}{Teh, Y.W.},
  \bibinfo{editor}{Titterington, M.} (Eds.), \bibinfo{booktitle}{Proceedings of
  the Thirteenth International Conference on Artificial Intelligence and
  Statistics}, \bibinfo{publisher}{PMLR}. pp. \bibinfo{pages}{249--256}.
\newblock \URLprefix \url{https://proceedings.mlr.press/v9/glorot10a.html}.
%Type = Misc
\bibitem[{Goyal et~al.(2017)Goyal, Dollár, Girshick, Noordhuis, Wesolowski,
  Kyrola, Tulloch, Jia and He}]{goyal_2017}
\bibinfo{author}{Goyal, P.}, \bibinfo{author}{Dollár, P.},
  \bibinfo{author}{Girshick, R.}, \bibinfo{author}{Noordhuis, P.},
  \bibinfo{author}{Wesolowski, L.}, \bibinfo{author}{Kyrola, A.},
  \bibinfo{author}{Tulloch, A.}, \bibinfo{author}{Jia, Y.},
  \bibinfo{author}{He, K.}, \bibinfo{year}{2017}.
\newblock \bibinfo{title}{Accurate, large minibatch sgd: Training imagenet in 1
  hour}.
\newblock \DOIprefix\doi{10.48550/ARXIV.1706.02677},
  \href{http://arxiv.org/abs/1706.02677}{{\tt arXiv:1706.02677}}.
%Type = Inproceedings
\bibitem[{He et~al.(2015)He, Zhang, Ren and Sun}]{10.1109/ICCV.2015.123}
\bibinfo{author}{He, K.}, \bibinfo{author}{Zhang, X.}, \bibinfo{author}{Ren,
  S.}, \bibinfo{author}{Sun, J.}, \bibinfo{year}{2015}.
\newblock \bibinfo{title}{Delving deep into rectifiers: Surpassing human-level
  performance on imagenet classification}, in: \bibinfo{booktitle}{Proceedings
  of the 2015 IEEE International Conference on Computer Vision (ICCV)},
  \bibinfo{publisher}{IEEE}. p. \bibinfo{pages}{1026–1034}.
\newblock \DOIprefix\doi{10.1109/ICCV.2015.123}.
%Type = Inproceedings
\bibitem[{He et~al.(2016)He, Zhang, Ren and Sun}]{he2016}
\bibinfo{author}{He, K.}, \bibinfo{author}{Zhang, X.}, \bibinfo{author}{Ren,
  S.}, \bibinfo{author}{Sun, J.}, \bibinfo{year}{2016}.
\newblock \bibinfo{title}{Deep residual learning for image recognition}, in:
  \bibinfo{booktitle}{2016 IEEE Conference on Computer Vision and Pattern
  Recognition (CVPR)}, pp. \bibinfo{pages}{770--778}.
\newblock \DOIprefix\doi{10.1109/CVPR.2016.90}.
%Type = Article
\bibitem[{Hochreiter and Schmidhuber(1997)}]{flat_minima}
\bibinfo{author}{Hochreiter, S.}, \bibinfo{author}{Schmidhuber, J.},
  \bibinfo{year}{1997}.
\newblock \bibinfo{title}{Flat minima}.
\newblock \bibinfo{journal}{Neural Comput.} \bibinfo{volume}{9},
  \bibinfo{pages}{1–42}.
\newblock \DOIprefix\doi{10.1162/neco.1997.9.1.1}.
%Type = Misc
\bibitem[{Inanc et~al.(2023a)Inanc, Sarma, Aach and Lintermann}]{ai4hpc}
\bibinfo{author}{Inanc, E.}, \bibinfo{author}{Sarma, R.},
  \bibinfo{author}{Aach, M.}, \bibinfo{author}{Lintermann, A.},
  \bibinfo{year}{2023}a.
\newblock \bibinfo{title}{{AI4HPC}}.
\newblock \DOIprefix\doi{10.5281/zenodo.7705417}.
%Type = Inproceedings
\bibitem[{Inanc et~al.(2023b)Inanc, Sarma, Aach, Sedona and
  Lintermann}]{inanc2023library}
\bibinfo{author}{Inanc, E.}, \bibinfo{author}{Sarma, R.},
  \bibinfo{author}{Aach, M.}, \bibinfo{author}{Sedona, R.},
  \bibinfo{author}{Lintermann, A.}, \bibinfo{year}{2023}b.
\newblock \bibinfo{title}{{AI4HPC}: Library to train {AI} models on {HPC}
  systems using {CFD} datasets}, in: \bibinfo{booktitle}{Workshop on Advancing
  Neural Network Training: Computational Efficiency, Scalability, and Resource
  Optimization (WANT@NeurIPS 2023)}.
\newblock \URLprefix \url{https://openreview.net/pdf?id=zQTa2XdPnP}.
%Type = Misc
\bibitem[{Jaderberg et~al.(2017)Jaderberg, Dalibard, Osindero, Czarnecki,
  Donahue, Razavi, Vinyals, Green, Dunning, Simonyan, Fernando and
  Kavukcuoglu}]{jaderberg2017population}
\bibinfo{author}{Jaderberg, M.}, \bibinfo{author}{Dalibard, V.},
  \bibinfo{author}{Osindero, S.}, \bibinfo{author}{Czarnecki, W.M.},
  \bibinfo{author}{Donahue, J.}, \bibinfo{author}{Razavi, A.},
  \bibinfo{author}{Vinyals, O.}, \bibinfo{author}{Green, T.},
  \bibinfo{author}{Dunning, I.}, \bibinfo{author}{Simonyan, K.},
  \bibinfo{author}{Fernando, C.}, \bibinfo{author}{Kavukcuoglu, K.},
  \bibinfo{year}{2017}.
\newblock \bibinfo{title}{Population based training of neural networks}.
\newblock \href{http://arxiv.org/abs/1711.09846}{{\tt arXiv:1711.09846}}.
%Type = Inproceedings
\bibitem[{Jamieson and Talwalkar(2016)}]{pmlr-v51-jamieson16}
\bibinfo{author}{Jamieson, K.}, \bibinfo{author}{Talwalkar, A.},
  \bibinfo{year}{2016}.
\newblock \bibinfo{title}{Non-stochastic best arm identification and
  hyperparameter optimization}, in: \bibinfo{editor}{Gretton, A.},
  \bibinfo{editor}{Robert, C.C.} (Eds.), \bibinfo{booktitle}{Proceedings of the
  19th International Conference on Artificial Intelligence and Statistics},
  \bibinfo{publisher}{PMLR}. pp. \bibinfo{pages}{240--248}.
\newblock \URLprefix \url{https://proceedings.mlr.press/v51/jamieson16.html}.
%Type = Inproceedings
\bibitem[{Jiang and Balaprakash(2020)}]{molecular_prediction}
\bibinfo{author}{Jiang, S.}, \bibinfo{author}{Balaprakash, P.},
  \bibinfo{year}{2020}.
\newblock \bibinfo{title}{Graph neural network architecture search for
  molecular property prediction}, in: \bibinfo{booktitle}{2020 IEEE
  International Conference on Big Data (Big Data)}, \bibinfo{publisher}{IEEE}.
  pp. \bibinfo{pages}{1346--1353}.
\newblock \DOIprefix\doi{10.1109/BigData50022.2020.9378060}.
%Type = Article
\bibitem[{{J\"{u}lich Supercomputing Centre}(2021)}]{JUWELS}
\bibinfo{author}{{J\"{u}lich Supercomputing Centre}}, \bibinfo{year}{2021}.
\newblock \bibinfo{title}{{JUWELS Cluster and Booster: Exascale Pathfinder with
  Modular Supercomputing Architecture at Juelich Supercomputing Centre}}.
\newblock \bibinfo{journal}{Journal of large-scale research facilities JLSRF}
  \bibinfo{volume}{7}.
\newblock \DOIprefix\doi{10.17815/jlsrf-7-183}.
%Type = Article
\bibitem[{{Jülich Supercomputing Centre}(2021)}]{juelich2021jureca}
\bibinfo{author}{{Jülich Supercomputing Centre}}, \bibinfo{year}{2021}.
\newblock \bibinfo{title}{{JURECA}: Data centric and booster modules
  implementing the modular supercomputing architecture at {Jülich
  Supercomputing Centre}}.
\newblock \bibinfo{journal}{Journal of large-scale research facilities JLSRF}
  \bibinfo{volume}{7}.
\newblock \DOIprefix\doi{10.17815/jlsrf-7-182}.
%Type = Article
\bibitem[{Kandasamy et~al.(2020)Kandasamy, Vysyaraju, Neiswanger, Paria,
  Collins, Schneider, Poczos and Xing}]{dragonfly}
\bibinfo{author}{Kandasamy, K.}, \bibinfo{author}{Vysyaraju, K.R.},
  \bibinfo{author}{Neiswanger, W.}, \bibinfo{author}{Paria, B.},
  \bibinfo{author}{Collins, C.R.}, \bibinfo{author}{Schneider, J.},
  \bibinfo{author}{Poczos, B.}, \bibinfo{author}{Xing, E.P.},
  \bibinfo{year}{2020}.
\newblock \bibinfo{title}{Tuning hyperparameters without grad students:
  Scalable and robust bayesian optimisation with dragonfly}.
\newblock \bibinfo{journal}{Journal of Machine Learning Research}
  \bibinfo{volume}{21}, \bibinfo{pages}{1--27}.
\newblock \URLprefix \url{http://jmlr.org/papers/v21/18-223.html}.
%Type = Misc
\bibitem[{Keskar et~al.(2017)Keskar, Mudigere, Nocedal, Smelyanskiy and
  Tang}]{keskar2017largebatch}
\bibinfo{author}{Keskar, N.S.}, \bibinfo{author}{Mudigere, D.},
  \bibinfo{author}{Nocedal, J.}, \bibinfo{author}{Smelyanskiy, M.},
  \bibinfo{author}{Tang, P.T.P.}, \bibinfo{year}{2017}.
\newblock \bibinfo{title}{On large-batch training for deep learning:
  Generalization gap and sharp minima}.
\newblock \DOIprefix\doi{10.48550/arXiv.1609.04836},
  \href{http://arxiv.org/abs/1609.04836}{{\tt arXiv:1609.04836}}.
%Type = Inproceedings
\bibitem[{Kesselheim et~al.(2021)Kesselheim, Herten, Krajsek, Ebert, Jitsev,
  Cherti, Langguth, Gong, Stadtler, Mozaffari, Cavallaro, Sedona, Schug,
  Strube, Kamath, Schultz, Riedel and Lippert}]{juwels_booster}
\bibinfo{author}{Kesselheim, S.}, \bibinfo{author}{Herten, A.},
  \bibinfo{author}{Krajsek, K.}, \bibinfo{author}{Ebert, J.},
  \bibinfo{author}{Jitsev, J.}, \bibinfo{author}{Cherti, M.},
  \bibinfo{author}{Langguth, M.}, \bibinfo{author}{Gong, B.},
  \bibinfo{author}{Stadtler, S.}, \bibinfo{author}{Mozaffari, A.},
  \bibinfo{author}{Cavallaro, G.}, \bibinfo{author}{Sedona, R.},
  \bibinfo{author}{Schug, A.}, \bibinfo{author}{Strube, A.},
  \bibinfo{author}{Kamath, R.}, \bibinfo{author}{Schultz, M.G.},
  \bibinfo{author}{Riedel, M.}, \bibinfo{author}{Lippert, T.},
  \bibinfo{year}{2021}.
\newblock \bibinfo{title}{{JUWELS} booster -- a supercomputer for large-scale
  {AI} research}, in: \bibinfo{editor}{Jagode, H.}, \bibinfo{editor}{Anzt, H.},
  \bibinfo{editor}{Ltaief, H.}, \bibinfo{editor}{Luszczek, P.} (Eds.),
  \bibinfo{booktitle}{High Performance Computing},
  \bibinfo{publisher}{Springer}. pp. \bibinfo{pages}{453--468}.
\newblock \DOIprefix\doi{10.1007/978-3-030-90539-2_31}.
%Type = Misc
\bibitem[{Krizhevsky(2014)}]{krizhevsky2014weird}
\bibinfo{author}{Krizhevsky, A.}, \bibinfo{year}{2014}.
\newblock \bibinfo{title}{One weird trick for parallelizing convolutional
  neural networks}.
\newblock \href{http://arxiv.org/abs/1404.5997}{{\tt arXiv:1404.5997}}.
%Type = Article
\bibitem[{Li et~al.(2018)Li, Jamieson, DeSalvo, Rostamizadeh and
  Talwalkar}]{hyperband}
\bibinfo{author}{Li, L.}, \bibinfo{author}{Jamieson, K.},
  \bibinfo{author}{DeSalvo, G.}, \bibinfo{author}{Rostamizadeh, A.},
  \bibinfo{author}{Talwalkar, A.}, \bibinfo{year}{2018}.
\newblock \bibinfo{title}{Hyperband: A novel bandit-based approach to
  hyperparameter optimization}.
\newblock \bibinfo{journal}{Journal of Machine Learning Research}
  \bibinfo{volume}{18}, \bibinfo{pages}{1--52}.
\newblock \URLprefix \url{https://jmlr.org/papers/v18/16-558.html}.
%Type = Inproceedings
\bibitem[{Li et~al.(2020a)Li, Jamieson, Rostamizadeh, Gonina, Ben-tzur, Hardt,
  Recht and Talwalkar}]{ASHA}
\bibinfo{author}{Li, L.}, \bibinfo{author}{Jamieson, K.},
  \bibinfo{author}{Rostamizadeh, A.}, \bibinfo{author}{Gonina, E.},
  \bibinfo{author}{Ben-tzur, J.}, \bibinfo{author}{Hardt, M.},
  \bibinfo{author}{Recht, B.}, \bibinfo{author}{Talwalkar, A.},
  \bibinfo{year}{2020}a.
\newblock \bibinfo{title}{A system for massively parallel hyperparameter
  tuning}, in: \bibinfo{editor}{Dhillon, I.}, \bibinfo{editor}{Papailiopoulos,
  D.}, \bibinfo{editor}{Sze, V.} (Eds.), \bibinfo{booktitle}{Proceedings of
  Machine Learning and Systems 2 (MLSys 2020)}, pp. \bibinfo{pages}{230--246}.
\newblock \URLprefix
  \url{https://proceedings.mlsys.org/paper_files/paper/2020/hash/a06f20b349c6cf09a6b171c71b88bbfc-Abstract.html}.
%Type = Article
\bibitem[{Li et~al.(2020b)Li, Zhao, Varma, Salpekar, Noordhuis, Li, Paszke,
  Smith, Vaughan, Damania and Chintala}]{pytorch_ddp}
\bibinfo{author}{Li, S.}, \bibinfo{author}{Zhao, Y.}, \bibinfo{author}{Varma,
  R.}, \bibinfo{author}{Salpekar, O.}, \bibinfo{author}{Noordhuis, P.},
  \bibinfo{author}{Li, T.}, \bibinfo{author}{Paszke, A.},
  \bibinfo{author}{Smith, J.}, \bibinfo{author}{Vaughan, B.},
  \bibinfo{author}{Damania, P.}, \bibinfo{author}{Chintala, S.},
  \bibinfo{year}{2020}b.
\newblock \bibinfo{title}{{PyTorch} distributed: Experiences on accelerating
  data parallel training}.
\newblock \bibinfo{journal}{Proceedings of Very Large Data Base Endowment}
  \bibinfo{volume}{13}, \bibinfo{pages}{3005–3018}.
\newblock \DOIprefix\doi{10.14778/3415478.3415530}.
%Type = Inproceedings
\bibitem[{Liaw et~al.(2019)Liaw, Bhardwaj, Dunlap, Zou, Gonzalez, Stoica and
  Tumanov}]{hyper_sched}
\bibinfo{author}{Liaw, R.}, \bibinfo{author}{Bhardwaj, R.},
  \bibinfo{author}{Dunlap, L.}, \bibinfo{author}{Zou, Y.},
  \bibinfo{author}{Gonzalez, J.E.}, \bibinfo{author}{Stoica, I.},
  \bibinfo{author}{Tumanov, A.}, \bibinfo{year}{2019}.
\newblock \bibinfo{title}{{HyperSched}: Dynamic resource reallocation for model
  development on a deadline}, in: \bibinfo{booktitle}{Proceedings of the ACM
  Symposium on Cloud Computing}, \bibinfo{publisher}{ACM}. p.
  \bibinfo{pages}{61–73}.
\newblock \DOIprefix\doi{10.1145/3357223.3362719}.
%Type = Misc
\bibitem[{Liaw et~al.(2018)Liaw, Liang, Nishihara, Moritz, Gonzalez and
  Stoica}]{ray_tune}
\bibinfo{author}{Liaw, R.}, \bibinfo{author}{Liang, E.},
  \bibinfo{author}{Nishihara, R.}, \bibinfo{author}{Moritz, P.},
  \bibinfo{author}{Gonzalez, J.E.}, \bibinfo{author}{Stoica, I.},
  \bibinfo{year}{2018}.
\newblock \bibinfo{title}{Tune: A research platform for distributed model
  selection and training}.
\newblock \href{http://arxiv.org/abs/1807.05118}{{\tt arXiv:1807.05118}}.
%Type = Article
\bibitem[{Lindauer et~al.(2022)Lindauer, Eggensperger, Feurer, Biedenkapp,
  Deng, Benjamins, Ruhkopf, Sass and Hutter}]{smac_3}
\bibinfo{author}{Lindauer, M.}, \bibinfo{author}{Eggensperger, K.},
  \bibinfo{author}{Feurer, M.}, \bibinfo{author}{Biedenkapp, A.},
  \bibinfo{author}{Deng, D.}, \bibinfo{author}{Benjamins, C.},
  \bibinfo{author}{Ruhkopf, T.}, \bibinfo{author}{Sass, R.},
  \bibinfo{author}{Hutter, F.}, \bibinfo{year}{2022}.
\newblock \bibinfo{title}{{SMAC3}: A versatile bayesian optimization package
  for hyperparameter optimization}.
\newblock \bibinfo{journal}{Journal of Machine Learning Research}
  \bibinfo{volume}{23}, \bibinfo{pages}{1--9}.
\newblock \URLprefix \url{http://jmlr.org/papers/v23/21-0888.html}.
%Type = Article
\bibitem[{Liu et~al.(2024)Liu, R{\"u}ttgers, Quercia, Egele, Pfaehler, Shende,
  Aach, Schr{\"o}der, Balaprakash and Lintermann}]{LIU2024474}
\bibinfo{author}{Liu, X.}, \bibinfo{author}{R{\"u}ttgers, M.},
  \bibinfo{author}{Quercia, A.}, \bibinfo{author}{Egele, R.},
  \bibinfo{author}{Pfaehler, E.}, \bibinfo{author}{Shende, R.},
  \bibinfo{author}{Aach, M.}, \bibinfo{author}{Schr{\"o}der, W.},
  \bibinfo{author}{Balaprakash, P.}, \bibinfo{author}{Lintermann, A.},
  \bibinfo{year}{2024}.
\newblock \bibinfo{title}{Refining computer tomography data with
  super-resolution networks to increase the accuracy of respiratory flow
  simulations}.
\newblock \bibinfo{journal}{Future Generation Computer Systems}
  \bibinfo{volume}{159}, \bibinfo{pages}{474--488}.
\newblock \DOIprefix\doi{https://doi.org/10.1016/j.future.2024.05.020}.
%Type = Inproceedings
\bibitem[{Loshchilov and Hutter(2017)}]{loshchilov2017sgdr}
\bibinfo{author}{Loshchilov, I.}, \bibinfo{author}{Hutter, F.},
  \bibinfo{year}{2017}.
\newblock \bibinfo{title}{{SGDR}: Stochastic gradient descent with warm
  restarts}, in: \bibinfo{booktitle}{International Conference on Learning
  Representations}.
\newblock \URLprefix \url{https://openreview.net/pdf?id=Skq89Scxx}.
%Type = Inproceedings
\bibitem[{Malladi et~al.(2022)Malladi, Lyu, Panigrahi and
  Arora}]{malladi2022on}
\bibinfo{author}{Malladi, S.}, \bibinfo{author}{Lyu, K.},
  \bibinfo{author}{Panigrahi, A.}, \bibinfo{author}{Arora, S.},
  \bibinfo{year}{2022}.
\newblock \bibinfo{title}{On the {SDE}s and scaling rules for adaptive gradient
  algorithms}, in: \bibinfo{editor}{Koyejo, S.}, \bibinfo{editor}{Mohamed, S.},
  \bibinfo{editor}{Agarwal, A.}, \bibinfo{editor}{Belgrave, D.},
  \bibinfo{editor}{Cho, K.}, \bibinfo{editor}{Oh, A.} (Eds.),
  \bibinfo{booktitle}{Proceedings of the 36th International Conference on
  Neural Information Processing Systems}, \bibinfo{publisher}{Curran
  Associates, Inc.}
\newblock \URLprefix
  \url{https://proceedings.neurips.cc/paper_files/paper/2022/file/32ac710102f0620d0f28d5d05a44fe08-Paper-Conference.pdf}.
%Type = Article
\bibitem[{Martinez-Cantin(2014)}]{bayes_opt}
\bibinfo{author}{Martinez-Cantin, R.}, \bibinfo{year}{2014}.
\newblock \bibinfo{title}{Bayesopt: A bayesian optimization library for
  nonlinear optimization, experimental design and bandits}.
\newblock \bibinfo{journal}{Journal of Machine Learning Research}
  \bibinfo{volume}{15}, \bibinfo{pages}{3915--3919}.
\newblock \URLprefix \url{http://jmlr.org/papers/v15/martinezcantin14a.html}.
%Type = Inproceedings
\bibitem[{Mattson et~al.(2020)Mattson, Cheng, Diamos, Coleman, Micikevicius,
  Patterson, Tang, Wei, Bailis, Bittorf, Brooks, Chen, Dutta, Gupta, Hazelwood,
  Hock, Huang, Kang, Kanter, Kumar, Liao, Narayanan, Oguntebi, Pekhimenko,
  Pentecost, Janapa~Reddi, Robie, St~John, Wu, Xu, Young and
  Zaharia}]{MLSYS2020_411e39b1}
\bibinfo{author}{Mattson, P.}, \bibinfo{author}{Cheng, C.},
  \bibinfo{author}{Diamos, G.}, \bibinfo{author}{Coleman, C.},
  \bibinfo{author}{Micikevicius, P.}, \bibinfo{author}{Patterson, D.},
  \bibinfo{author}{Tang, H.}, \bibinfo{author}{Wei, G.Y.},
  \bibinfo{author}{Bailis, P.}, \bibinfo{author}{Bittorf, V.},
  \bibinfo{author}{Brooks, D.}, \bibinfo{author}{Chen, D.},
  \bibinfo{author}{Dutta, D.}, \bibinfo{author}{Gupta, U.},
  \bibinfo{author}{Hazelwood, K.}, \bibinfo{author}{Hock, A.},
  \bibinfo{author}{Huang, X.}, \bibinfo{author}{Kang, D.},
  \bibinfo{author}{Kanter, D.}, \bibinfo{author}{Kumar, N.},
  \bibinfo{author}{Liao, J.}, \bibinfo{author}{Narayanan, D.},
  \bibinfo{author}{Oguntebi, T.}, \bibinfo{author}{Pekhimenko, G.},
  \bibinfo{author}{Pentecost, L.}, \bibinfo{author}{Janapa~Reddi, V.},
  \bibinfo{author}{Robie, T.}, \bibinfo{author}{St~John, T.},
  \bibinfo{author}{Wu, C.J.}, \bibinfo{author}{Xu, L.}, \bibinfo{author}{Young,
  C.}, \bibinfo{author}{Zaharia, M.}, \bibinfo{year}{2020}.
\newblock \bibinfo{title}{{MLPerf} training benchmark}, in:
  \bibinfo{editor}{Dhillon, I.}, \bibinfo{editor}{Papailiopoulos, D.},
  \bibinfo{editor}{Sze, V.} (Eds.), \bibinfo{booktitle}{Proceedings of Machine
  Learning and Systems}, pp. \bibinfo{pages}{336--349}.
\newblock \URLprefix
  \url{https://proceedings.mlsys.org/paper_files/paper/2020/file/411e39b117e885341f25efb8912945f7-Paper.pdf}.
%Type = Misc
\bibitem[{McCandlish et~al.(2018)McCandlish, Kaplan, Amodei and
  Team}]{mccandlish2018empirical}
\bibinfo{author}{McCandlish, S.}, \bibinfo{author}{Kaplan, J.},
  \bibinfo{author}{Amodei, D.}, \bibinfo{author}{Team, O.D.},
  \bibinfo{year}{2018}.
\newblock \bibinfo{title}{An empirical model of large-batch training}.
\newblock \href{http://arxiv.org/abs/1812.06162}{{\tt arXiv:1812.06162}}.
%Type = Inproceedings
\bibitem[{Misra et~al.(2021)Misra, Liaw, Dunlap, Bhardwaj, Kandasamy, Gonzalez,
  Stoica and Tumanov}]{rubberband}
\bibinfo{author}{Misra, U.}, \bibinfo{author}{Liaw, R.},
  \bibinfo{author}{Dunlap, L.}, \bibinfo{author}{Bhardwaj, R.},
  \bibinfo{author}{Kandasamy, K.}, \bibinfo{author}{Gonzalez, J.E.},
  \bibinfo{author}{Stoica, I.}, \bibinfo{author}{Tumanov, A.},
  \bibinfo{year}{2021}.
\newblock \bibinfo{title}{Rubberband: cloud-based hyperparameter tuning}, in:
  \bibinfo{booktitle}{Proceedings of the Sixteenth European Conference on
  Computer Systems}, \bibinfo{publisher}{ACM}. p. \bibinfo{pages}{327–342}.
\newblock \DOIprefix\doi{10.1145/3447786.3456245}.
%Type = Article
\bibitem[{Pata et~al.(2024)Pata, Wulff, Mokhtar, Southwick, Zhang, Girone and
  Duarte}]{cern_mlpf_data}
\bibinfo{author}{Pata, J.}, \bibinfo{author}{Wulff, E.},
  \bibinfo{author}{Mokhtar, F.}, \bibinfo{author}{Southwick, D.},
  \bibinfo{author}{Zhang, M.}, \bibinfo{author}{Girone, M.},
  \bibinfo{author}{Duarte, J.}, \bibinfo{year}{2024}.
\newblock \bibinfo{title}{Improved particle-flow event reconstruction with
  scalable neural networks for current and future particle detectors}.
\newblock \bibinfo{journal}{Communications Physics} \bibinfo{volume}{7},
  \bibinfo{pages}{124}.
\newblock \DOIprefix\doi{10.1038/s42005-024-01599-5}.
%Type = Inproceedings
\bibitem[{Qiao et~al.(2021)Qiao, Choe, Subramanya, Neiswanger, Ho, Zhang,
  Ganger and Xing}]{pollux}
\bibinfo{author}{Qiao, A.}, \bibinfo{author}{Choe, S.K.},
  \bibinfo{author}{Subramanya, S.J.}, \bibinfo{author}{Neiswanger, W.},
  \bibinfo{author}{Ho, Q.}, \bibinfo{author}{Zhang, H.},
  \bibinfo{author}{Ganger, G.R.}, \bibinfo{author}{Xing, E.P.},
  \bibinfo{year}{2021}.
\newblock \bibinfo{title}{Pollux: Co-adaptive cluster scheduling for
  goodput-optimized deep learning}, in: \bibinfo{booktitle}{15th {USENIX}
  Symposium on Operating Systems Design and Implementation ({OSDI} 21)},
  \bibinfo{publisher}{{USENIX} Association}. pp. \bibinfo{pages}{1--18}.
\newblock \URLprefix
  \url{https://www.usenix.org/conference/osdi21/presentation/qiao}.
%Type = Article
\bibitem[{Russakovsky et~al.(2015)Russakovsky, Deng, Su, Krause, Satheesh, Ma,
  Huang, Karpathy, Khosla, Bernstein, Berg and Fei-Fei}]{imagenet}
\bibinfo{author}{Russakovsky, O.}, \bibinfo{author}{Deng, J.},
  \bibinfo{author}{Su, H.}, \bibinfo{author}{Krause, J.},
  \bibinfo{author}{Satheesh, S.}, \bibinfo{author}{Ma, S.},
  \bibinfo{author}{Huang, Z.}, \bibinfo{author}{Karpathy, A.},
  \bibinfo{author}{Khosla, A.}, \bibinfo{author}{Bernstein, M.},
  \bibinfo{author}{Berg, A.C.}, \bibinfo{author}{Fei-Fei, L.},
  \bibinfo{year}{2015}.
\newblock \bibinfo{title}{Imagenet large scale visual recognition challenge}.
\newblock \bibinfo{journal}{International Journal of Computer Vision}
  \bibinfo{volume}{115}, \bibinfo{pages}{211--252}.
\newblock \DOIprefix\doi{10.1007/s11263-015-0816-y}.
%Type = Inproceedings
\bibitem[{Smith et~al.(2018)Smith, Kindermans and Le}]{smith2018dont}
\bibinfo{author}{Smith, S.L.}, \bibinfo{author}{Kindermans, P.J.},
  \bibinfo{author}{Le, Q.V.}, \bibinfo{year}{2018}.
\newblock \bibinfo{title}{Don't decay the learning rate, increase the batch
  size}, in: \bibinfo{booktitle}{International Conference on Learning
  Representations}.
\newblock \URLprefix \url{https://openreview.net/pdf?id=B1Yy1BxCZ}.
%Type = Inproceedings
\bibitem[{Sumbul et~al.(2019)Sumbul, Charfuelan, Demir and
  Markl}]{big_earth_net}
\bibinfo{author}{Sumbul, G.}, \bibinfo{author}{Charfuelan, M.},
  \bibinfo{author}{Demir, B.}, \bibinfo{author}{Markl, V.},
  \bibinfo{year}{2019}.
\newblock \bibinfo{title}{Bigearthnet: A large-scale benchmark archive for
  remote sensing image understanding}, in: \bibinfo{booktitle}{IGARSS 2019 -
  2019 IEEE International Geoscience and Remote Sensing Symposium},
  \bibinfo{publisher}{IEEE}. pp. \bibinfo{pages}{5901--5904}.
\newblock \DOIprefix\doi{10.1109/IGARSS.2019.8900532}.
%Type = Inproceedings
\bibitem[{Taubert et~al.(2023)Taubert, Weiel, Coquelin, Farshian, Debus, Schug,
  Streit and G{\"o}tz}]{propulate}
\bibinfo{author}{Taubert, O.}, \bibinfo{author}{Weiel, M.},
  \bibinfo{author}{Coquelin, D.}, \bibinfo{author}{Farshian, A.},
  \bibinfo{author}{Debus, C.}, \bibinfo{author}{Schug, A.},
  \bibinfo{author}{Streit, A.}, \bibinfo{author}{G{\"o}tz, M.},
  \bibinfo{year}{2023}.
\newblock \bibinfo{title}{Massively parallel genetic optimization through
  asynchronous propagation of populations}, in: \bibinfo{editor}{Bhatele, A.},
  \bibinfo{editor}{Hammond, J.}, \bibinfo{editor}{Baboulin, M.},
  \bibinfo{editor}{Kruse, C.} (Eds.), \bibinfo{booktitle}{High Performance
  Computing. ISC High Performance 2023}, \bibinfo{publisher}{Springer}. pp.
  \bibinfo{pages}{106--124}.
\newblock \DOIprefix\doi{10.1007/978-3-031-32041-5_6}.
%Type = Misc
\bibitem[{Yang et~al.(2023)Yang, Guo, Xiong, Liu, Pan, Wang, Tong and
  Guo}]{yang2023swin3dpretrainedtransformerbackbone}
\bibinfo{author}{Yang, Y.Q.}, \bibinfo{author}{Guo, Y.X.},
  \bibinfo{author}{Xiong, J.Y.}, \bibinfo{author}{Liu, Y.},
  \bibinfo{author}{Pan, H.}, \bibinfo{author}{Wang, P.S.},
  \bibinfo{author}{Tong, X.}, \bibinfo{author}{Guo, B.}, \bibinfo{year}{2023}.
\newblock \bibinfo{title}{Swin3d: A pretrained transformer backbone for 3d
  indoor scene understanding}.
\newblock \href{http://arxiv.org/abs/2304.06906}{{\tt arXiv:2304.06906}}.

\end{thebibliography}





% \end{thebibliography}

\appendix
\newpage
\quad
\newpage
\section{\ac{RASDA} GitHub Repository}
% \begin{table}[hb]
% \caption{Content of the \ac{RASDA} GitHub Repository}
% \label{tab:git_repo_table}
% \centering
% \begin{tabular}{|l|l|}
% \hline
% \textbf{File} & \textbf{Description} \\
% \hline
% \texttt{startscript.sh} & Shell script to launch the HPO process via Ray on an HPC system \\
% \texttt{build\_ray\_env.sh} & Shell script to set up the Ray environment on an HPC system \\
% \texttt{adaptive\_ray.py} & Main script to configure the schedulers and launch the HPO trials \\
% \texttt{res\_allocate.py} & Core resource allocation logic for RASDA \\
% \texttt{cases/ImagenetTrainLoopDALI.py} & Script containing the training logic of the \ac{CV} application example \\
% \hline
% \end{tabular}
% \end{table}
\begin{table}[!ht]
\caption{Content of the \ac{RASDA} GitHub Repository}
\label{tab:git_repo_table}
\centering
\begin{tabular}{|l|l|}
\hline
\textcolor{black}{\textbf{File}} & \textcolor{black}{\textbf{Description}} \\
\hline
\textcolor{black}{\texttt{startscript.sh}} & \textcolor{black}{Shell script to launch the HPO process via Ray on an HPC system} \\
\textcolor{black}{\texttt{build\_ray\_env.sh}} & \textcolor{black}{Shell script to set up the Ray environment on an HPC system} \\
\textcolor{black}{\texttt{adaptive\_ray.py}} & \textcolor{black}{Main script to configure the schedulers and launch the HPO trials} \\
\textcolor{black}{\texttt{res\_allocate.py}} & \textcolor{black}{Core resource allocation logic for RASDA} \\
\textcolor{black}{\texttt{cases/ImagenetTrainLoopDALI.py}} & \textcolor{black}{Script containing the training logic of the \ac{CV} application example} \\
\hline
\end{tabular}
\end{table}

\end{document}